\documentclass[10pt,twocolumn,letterpaper]{article}

\usepackage{cvpr}              

\definecolor{cvprblue}{rgb}{0.21,0.49,0.74}
\usepackage[pagebackref,breaklinks,colorlinks,allcolors=cvprblue]{hyperref}
\usepackage{adjustbox}

\usepackage{utfsym}

\usepackage{multirow,algorithm}

\usepackage[utf8]{inputenc} 
\usepackage[T1]{fontenc}    
\usepackage{booktabs}       
\usepackage{amsfonts}       
\usepackage{nicefrac}       
\usepackage{microtype}      
\usepackage[dvipsnames]{xcolor}
\usepackage{colortbl}
\usepackage{enumitem}
\usepackage{algpseudocode}
\usepackage{times}
\usepackage{epsfig}
\usepackage{amssymb}
\usepackage{caption}
\usepackage{subcaption}
\usepackage{tabularx}
\usepackage{rotating}
\usepackage{etoolbox}
\usepackage[normalem]{ulem}
\usepackage{lipsum}
\usepackage{stmaryrd}
\usepackage{stackengine}
\usepackage{makecell}
\usepackage{pifont}
\usepackage{cancel}
\usepackage{wrapfig}
\usepackage{bm}

\usepackage{tikz}

\definecolor{barrier}{RGB}{112,128,144}
\definecolor{bicycle}{RGB}{220,20,60}
\definecolor{bus}{RGB}{255, 127, 80}
\definecolor{car}{RGB}{255, 158, 0}
\definecolor{const. veh.}{RGB}{233, 150, 70}
\definecolor{motorcycle}{RGB}{255,61,99}
\definecolor{pedestrian}{RGB}{0,0,230}
\definecolor{traffic cone}{RGB}{47,79,79}
\definecolor{trailer}{RGB}{255,140,0}
\definecolor{truck}{RGB}{255,99,71}
\definecolor{drive. suf.}{RGB}{0,207,191}
\definecolor{other flat}{RGB}{175,0,75}
\definecolor{sidewalk}{RGB}{75,0,75}
\definecolor{terrain}{RGB}{112,180,60}
\definecolor{manmade}{RGB}{222,184,135}
\definecolor{vegetation}{RGB}{0,175,0}
\definecolor{y}{HTML}{00994C}

\definecolor{other flat1}{RGB}{175,0,75}
\newcommand{\hi}[1]{\textbf{\textcolor{other flat1}{#1}}}

\def\paperID{} 
\def\confName{CVPR}
\def\confYear{2025}

\title{

\hspace{0.4cm} UniScene: Unified Occupancy-centric Driving Scene Generation 
}

\author{
\small
Bohan Li$^{1,2}$\thanks{Equal contribution}, Jiazhe Guo$^{3*}$, Hongsi Liu$^{2*}$, Yingshuang Zou$^{3*}$, Yikang Ding$^{4*}$, Xiwu Chen$^{5}$, Hu Zhu$^{2}$, Feiyang Tan$^{5}$, Chi Zhang$^{5}$, \\
\small
Tiancai Wang$^{4}$, Shuchang Zhou$^{4}$, Li Zhang$^{6}$, Xiaojuan Qi$^{7}$, Hao Zhao$^{3}$, Mu Yang$^{4}$, Wenjun Zeng$^{2}$, Xin Jin$^{2}$\thanks{Corresponding author}  \\ \vspace{-0.6em} \\
\small
$^{1}$Shanghai Jiao Tong University, $^{2}$Ningbo Institute of Digital Twin, Eastern Institute of Technology, China \\
\small
$^{3}$Tsinghua University, $^{4}$MEGVII Technology, $^{5}$Mach Drive, $^{6}$Fudan University, $^{7}$University of Hong Kong
}

\makeatletter 
\g@addto@macro\@maketitle{
\vspace{-1.0cm}
\begin{tikzpicture}[remember picture,overlay,shift={(current page.north west)}]
\node[anchor=north west, xshift=3.16cm, yshift=-3.15cm]{\scalebox{1}[1]{\includegraphics[width=0.8cm]{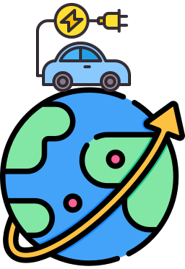}}};
\end{tikzpicture}

 \vspace{-10pt}
  \begin{figure}[H]
  \setlength{\linewidth}{\textwidth}
  \setlength{\hsize}{\textwidth}
  \centering
    \includegraphics[width=0.99\linewidth]{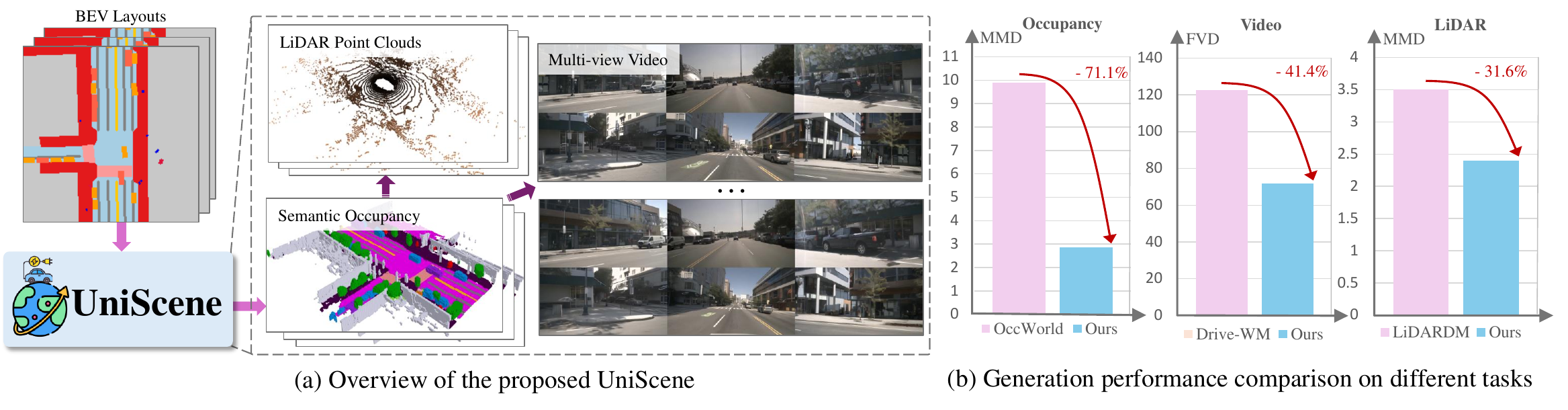}
     \vspace{-8pt}
  \caption{\textbf{(a) Overview of UniScene.} Given BEV layouts, UniScene facilitates versatile data generation, including semantic occupancy, multi-view video, and LiDAR point clouds, through an occupancy-centric hierarchical modeling approach. \textbf{(b) Performance comparison on different generation tasks.} UniScene delivers substantial improvements over SOTA methods in video, LiDAR, and occupancy generation. 
  }
 \vspace{-5pt}
  \label{fig_teaser1}
  \end{figure}
}

\begin{document}
\maketitle

\vspace{-0.8cm}
\begin{abstract}
Generating high-fidelity, controllable, and annotated training data is critical for autonomous driving. Existing methods typically generate a single data form directly from a coarse scene layout, which not only fails to output rich data forms required for diverse downstream tasks but also struggles to model the direct layout-to-data distribution.
In this paper, we introduce UniScene, the first unified framework for generating three key data forms — semantic occupancy, video, and LiDAR — in driving scenes. UniScene employs a progressive generation process that decomposes the complex task of scene generation into two hierarchical steps: (a) first generating semantic occupancy from a customized scene layout as a meta scene representation rich in both semantic and geometric information, and then (b) conditioned on occupancy, generating video and LiDAR data, respectively, with two novel transfer strategies of Gaussian-based Joint Rendering and Prior-guided Sparse Modeling. This occupancy-centric approach reduces the generation burden, especially for intricate scenes, while providing detailed intermediate representations for the subsequent generation stages. Extensive experiments demonstrate that UniScene outperforms previous SOTAs in the occupancy, video, and LiDAR generation, which also indeed benefits downstream driving tasks. Project page: \url{https://arlo0o.github.io/uniscene/}.
\end{abstract}
\vspace{-0.8cm}

\section{Introduction}
\vspace{-0.1cm}
The generation of high-quality driving scenes is a promising approach for autonomous driving (AD), as it helps mitigate the high resource demands associated with real-world data collection and annotation~\cite{luo2021diffusion, rombach2022high, jiang2022conditional, li2024time}. Recent advancements in generative models, particularly diffusion models~\cite{luo2021diffusion, rombach2022high, jiang2022conditional, li2024time}, have made it possible to generate realistic synthetic data~\cite{yang2023bevcontrol, swerdlow2024street, wang2023drivedreamer}, facilitating the training of downstream tasks.
Existing methods~\cite{wang2023drivedreamer, gao2023magicdrive, wang2023driving, wen2023panacea, zyrianov2024lidardm} typically use layout conditions derived from coarse geometric labels (\eg, BEV maps and 3D bounding boxes) as input to guide scene generation.
The resulting synthetic data are then leveraged to improve downstream tasks such as BEV segmentation~\cite{li2023open, wu2023datasetdm, li2024fairdiff} and 3D object detection~\cite{bowles2018gan, chen2023integrating, wang2024detdiffusion, he2022synthetic, moller2023prompt}.

Nevertheless, as depicted in Tab.~\ref{teaser_tab}, existing driving scene generation models predominantly focus on generating data in a single format (\eg, RGB video)~\cite{wang2023drivedreamer,zhao2024drivedreamer,gao2023magicdrive,wang2023driving,wen2024panacea}, without fully exploring the potential of generating data across multiple formats. This limits their applicability for a wide range of downstream tasks that require diverse sensor data (\textit{i.e.}, RGB video, LiDAR) to ensure sufficient training for real-world scenarios~\cite{wang2023openoccupancy,liang2022bevfusion,li2022deepfusion,bai2022transfusion,berrio2021camera}.
Furthermore, previous methods attempt to capture the real-world distribution with a single-step layout-to-data modeling process given only coarse input conditions (\eg, BEV layouts or 3D boxes)~\cite{gao2023magicdrive,wang2023driving,wen2024panacea}. 
This direct learning strategy hinders the model's ability to capture the complex distributions inherent in real-world driving scenes (\eg, realistic geometry and appearance), often resulting in suboptimal performance, as illustrated in Fig.~\ref{fig_teaser1} (b).
To address this challenge, recent methods in data-driven realistic generation~\cite{koratana2019lit, men2020controllable, tewari2023diffusion} have sought to model complex distributions using intermediate representations as inductive biases, enabling the generation of high-quality results through hierarchical steps.

Thus, exploring an optimal intermediate representation for complex 3D generation tasks in autonomous driving is crucial for achieving high-quality outputs. Semantic occupancy, widely used in autonomous driving perception tasks, has recently been recognized as a superior scene representation due to its rich semantic and geometric information~\cite{tong2023scene, hu2023planning, wang2023openoccupancy}. Building on this, recent advancements in volumetric generation~\cite{zyrianov2022learning, ran2024towards, zyrianov2024lidardm, li2024time} highlight the significant potential of semantic occupancy, not only for depicting driving environments with enhanced 3D structural details but also for enabling more accurate and diverse scene generation. Compared to traditional 2D representations, such as BEV maps~\cite{yang2023bevcontrol, swerdlow2024street, wang2023drivedreamer, zhao2024drivedreamer, gao2023magicdrive, wang2023driving, wen2024panacea}, 3D occupancy offers a richer and more detailed scene representation. Given these advantages, we argue that semantic occupancy is an ideal intermediate representation for decomposing complex driving scene generation tasks. It captures both semantic and geometric information, facilitating the generation of diverse data formats (\eg, RGB video and LiDAR) while enhancing the flexibility and accuracy of the generation process.

\begin{table}[!t]
\vspace{-0pt}
\begin{center}
\scriptsize
\renewcommand\tabcolsep{5.6pt}
\centering
\resizebox{1.00\linewidth}{!}{
\begin{tabular}{l|cc|c|c}
\toprule Method & Multi-view & Video & LiDAR & Occupancy \\ \midrule
BEVGen~\cite{swerdlow2024streetview} & \textcolor{ForestGreen}{\usym{2713}} & \textcolor{red}{\usym{2717}} & \textcolor{red}{\usym{2717}} & \textcolor{red}{\usym{2717}}   \\
BEVControl~\cite{yang2023bevcontrol} & \textcolor{ForestGreen}{\usym{2713}} & \textcolor{red}{\usym{2717}}& \textcolor{red}{\usym{2717}} & \textcolor{red}{\usym{2717}}\\
DriveDreamer~\cite{zhao2024drivedreamer} & \textcolor{red}{\usym{2717}} & \textcolor{ForestGreen}{\usym{2713}} & \textcolor{red}{\usym{2717}} &  \textcolor{red}{\usym{2717}} \\

Vista~\cite{gao2024vista} & \textcolor{red}{\usym{2717}} & \textcolor{ForestGreen}{\usym{2713}}& \textcolor{red}{\usym{2717}} &  \textcolor{red}{\usym{2717}}\\ 
MagicDrive~\cite{gao2023magicdrive} & \textcolor{ForestGreen}{\usym{2713}} & \textcolor{ForestGreen}{\usym{2713}} & \textcolor{red}{\usym{2717}} & \textcolor{red}{\usym{2717}} \\
Drive-WM~\cite{wang2023driving} & \textcolor{ForestGreen}{\usym{2713}} & \textcolor{ForestGreen}{\usym{2713}} & \textcolor{red}{\usym{2717}} & \textcolor{red}{\usym{2717}}  \\

   \midrule

LiDARGen~\cite{zyrianov2022learning}   & \textcolor{red}{\usym{2717}}& \textcolor{red}{\usym{2717}}& \textcolor{ForestGreen}{\usym{2713}} &  \textcolor{red}{\usym{2717}} \\
LiDARDiffusion~\cite{ran2024towards} &\textcolor{red}{\usym{2717}} &\textcolor{red}{\usym{2717}} & \textcolor{ForestGreen}{\usym{2713}} & \textcolor{red}{\usym{2717}} \\
LiDARDM~\cite{zyrianov2024lidardm} & \textcolor{red}{\usym{2717}} & \textcolor{red}{\usym{2717}} & \textcolor{ForestGreen}{\usym{2713}} & \textcolor{red}{\usym{2717}}  \\  \midrule

OccSora~\cite{wang2024occsora} &\textcolor{red}{\usym{2717}} &\textcolor{red}{\usym{2717}} & \textcolor{red}{\usym{2717}} & \textcolor{ForestGreen}{\usym{2713}}  \\ 
OccLLama~\cite{wei2024occllama} &\textcolor{red}{\usym{2717}} & \textcolor{red}{\usym{2717}} & \textcolor{red}{\usym{2717}} & \textcolor{ForestGreen}{\usym{2713}}  \\
OccWorld~\cite{zheng2023occworld}  & \textcolor{red}{\usym{2717}} & \textcolor{red}{\usym{2717}} & \textcolor{red}{\usym{2717}} & \textcolor{ForestGreen}{\usym{2713}}   \\ 
\midrule

\rowcolor{gray!10}UniScene (Ours) & \textcolor{ForestGreen}{\usym{2713}}&  \textcolor{ForestGreen}{\usym{2713}}& \textcolor{ForestGreen}{\usym{2713}}&  \textcolor{ForestGreen}{\usym{2713}}\\
\bottomrule
\end{tabular}
 }
 \vspace{-8pt}
\caption{\textbf{Comparison of generation forms} with existing works.}
\vspace{-28pt}
\label{teaser_tab}
\end{center}
\end{table}

To this end, we propose UniScene, a unified occupancy-centric framework designed for the versatile generation of semantic occupancy, video, and LiDAR data. As illustrated in Fig.~\ref{fig_teaser1} (a), UniScene adopts a decomposed learning paradigm and is structured hierarchically: it first generates 3D semantic occupancy from BEV scene layouts, and subsequently leverages this representation to facilitate the generation of video and LiDAR data. Specifically, in contrast to previous unconditional occupancy generation approaches~\cite{lee2023diffusion, lee2024semcity, liu2023pyramid}, we use customized BEV layout sequences as controllable inputs to generate semantic occupancy sequences with spatial-temporal consistency. Unlike single-step layout-to-data learning methods~\cite{gao2023magicdrive, wang2023driving,wen2024panacea,swerdlow2024streetview,zhao2024drivedreamer}, our approach utilizes the generated occupancy as an intermediate representation to guide the subsequent generation.
To bridge the representation gap and ensure high-fidelity generation of video and LiDAR data, we introduce two novel representation transfer strategies:
1). A geometry-semantics joint rendering strategy, utilizing Gaussian Splatting~\cite{kerbl20233d, zhou2024hugs}, to facilitate conditional video generation with detailed multi-view semantic and depth maps;
2). A prior-guided sparse modeling scheme for LiDAR data generation, which efficiently generates LiDAR points using occupancy-based priors. 
The contributions of our framework can be summarized as follows:\looseness=-1

\begin{itemize}

    \item We introduce UniScene, the first unified framework for versatile data generation in driving scenes. It jointly generates high-quality data across three formats: semantic occupancy, multi-view video, and LiDAR point clouds.
    
    \item We propose a decomposed conditional generation paradigm that progressively models complex driving scenes, effectively reducing the difficulty of generation. Fine-grained semantic occupancy is first generated as an intermediate representation, which then facilitates the subsequent generation of video and LiDAR data.
    
    \item To bridge the domain gap between occupancy and other data formats, we introduce two novel representation transfer strategies: one based on Gaussian Splatting rendering and the other leveraging a sparse modeling scheme.

    \item Extensive experiments across various generation tasks demonstrate that UniScene outperforms state-of-the-art methods in video, LiDAR, and occupancy generation. Moreover, the data generated by UniScene leads to significant enhancements in downstream tasks, including occupancy prediction, 3D detection, and BEV segmentation.
\end{itemize}

\begin{figure*}[!t]
    \centering
    \vspace{-25pt}
    \includegraphics[width=1.0\linewidth]{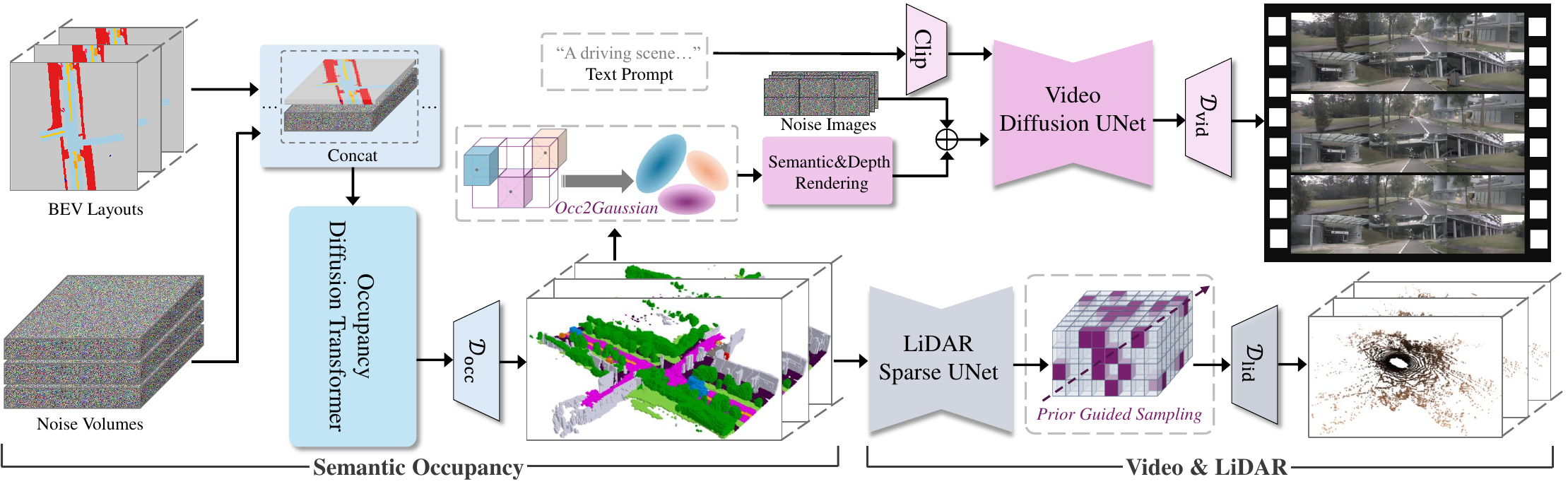}
    \vspace{-23pt}
    \caption{\textbf{Overall framework of the proposed method.}
    The joint generation process is organized into an occupancy-centric hierarchy: \textbf{I. Controllable Occupancy Generation.} The BEV layouts are concatenated with the noise volumes before being fed into the Occupancy Diffusion Transformer, and decoded with the Occupancy VAE Decoder $\mathcal{D}_\mathrm{occ}$. \textbf{II. Occupancy-based Video and LiDAR Generation.} 
     The occupancy is converted into 3D Gaussians and rendered into semantic and depth maps, which are processed with additional encoders as in ControlNet. The output is obtained from the Video VAE Decoder $\mathcal{D}_\mathrm{vid}$.
    For LiDAR generation, the occupancy is processed via a sparse UNet and sampled with the geometric prior guidance, which is sent to the LiDAR head $\mathcal{D}_\mathrm{lid}$ for generation.
    }
    \label{fig_overall}
    \vspace{-18pt}
\end{figure*}

\vspace{-5pt}
\section{Related Work}
\noindent\textbf{Semantic Occupancy Representation.}
Semantic occupancy has emerged as a prominent 3D scene representation in autonomous driving. Current research primarily focuses on Semantic Occupancy Prediction (SOP). MonoScene~\cite{cao2022monoscene} introduces a 3D SOP method based on a monocular camera. 
TPVFormer~\cite{huang2023tri} presents a Tri-Perspective View framework for SOP. 
VPD~\cite{li2024time} utilizes a generative diffusion model for 3D SOP. 
In the realm of occupancy forecasting, OccWorld~\cite{zheng2023occworld} predicts future occupancy states based on prior occupancy, while OccLlama~\cite{wei2024occllama} incorporates Large Language Models (LLMs) to assist in future occupancy prediction. However, research on occupancy generation, particularly for temporal 3D occupancy sequence generation, is still limited. Recently, OccSora~\cite{wang2024occsora} employs a Diffusion Transformer (DiT) to generate occupancy, but the quality of the generated results still lags behind the ground truth.

\noindent\textbf{Generation Models in Autonomous Driving.}
High-quality data is essential for training models in autonomous driving, which has led to a growing interest in driving scene generation tasks. One line of research employs Neural Radiance Fields (NeRF) and Gaussian Splatting (GS) techniques~\cite{wu2023mars, yan2024street, yariv2024diverse, xiangli2022bungeenerf} to synthesize novel perspectives, though these methods often suffer from limited scene diversity. With the rise of diffusion models, increasing attention has been given to generating driving images or videos, as seen in works like BEVGen~\cite{swerdlow2024streetview}, DriveDreamer~\cite{wang2023drivedreamer}, MagicDrive~\cite{gao2023magicdrive}, and Panacea~\cite{wen2023panacea}. Some approaches also integrate world-model concepts into the generation process, such as Drive-WM~\cite{wang2023driving}, WoVoGen~\cite{lu2023wovogen}, and Vista~\cite{gao2024vista}. In addition to generating images or videos, recent studies have explored the generation of LiDAR point clouds, including LidarDiffusion~\cite{ran2024towards} and LidarDM~\cite{zyrianov2024lidardm}. 
Nevertheless, these methods primarily focus on single-form generation, overlooking the complementary nature of multi-modal data. In contrast, our versatile framework generates multiple forms of high-quality data, including occupancy, video, and LiDAR.

\vspace{-5pt}
\section{Methodology}

In this section, we present UniScene, a unified framework designed to jointly generate three data forms: semantic occupancy, multi-view video, and LiDAR point cloud.

\noindent\textbf{Overview.}
As shown in Fig.~\ref{fig_overall}, we decompose the complex task of driving scene generation into an occupancy-centric hierarchy.
Specifically, given multi-frame BEV layouts as conditions, UniScene first generates the corresponding semantic occupancy sequence with an Occupancy Diffusion Transformer (Sec.~\ref{sec_occ}).
The generated occupancy then serves as conditional guidance for subsequent video and LiDAR generation.
For video generation, the occupancy is converted into 3D Gaussian primitives, which are then rendered into 2D semantic and depth maps to guide the video diffusion UNet (Sec.~\ref{sec_video}).
For LiDAR generation, we propose a sparse modeling approach that combines a LiDAR sparse UNet with a ray-based sparse sampling strategy, guided by occupancy priors, to effectively produce LiDAR points (Sec.~\ref{sec_lidar}).

\subsection{Controllable Semantic Occupancy Generation}\label{sec_occ}  

Generating controllable and temporally consistent semantic occupancy is crucial in UniScene, as subsequent video and LiDAR generation depend on it. To address this, we introduce the Occupancy Diffusion Transformer (DiT), which takes BEV layout sequences as input, allowing users to easily edit and generate the corresponding occupancy sequences.

\noindent\textbf{Temporal-aware Occupancy VAE.}
The occupancy VAE is designed to compress occupancy data into a latent space for computational efficiency. Unlike existing VQVAE-based approaches that rely on discrete tokenizers~\cite{zheng2023occworld,wei2024occllama}, our method employs a VAE to encode occupancy sequences into a continuous latent space. It facilitates better preservation of spatial details, particularly under high compression ratios. Experimental evaluations are provided in Sec.\ref{exp_occ} and Tab.\ref{tab_occ_rec}.

Specifically, temporal information is only considered during the VAE decoding process, enabling greater flexibility as in~\cite{blattmann2023align}. In the VAE encoder, we transform the 3D occupancy $\mathbf{O} \in \mathbb{R}^{H \times W \times D}$ into a BEV representation $\mathbf{\hat{O}} \in \mathbb{R}^{H \times W \times DC'}$ following~\cite{zheng2023occworld}, where $C'$ represents the dimension of the learnable class embedding. Subsequently, 2D convolutional layers and 2D axial attention layers are used to obtain a down-sampled continuous latent feature.
The VAE decoder reconstructs the temporal latent feature $\boldsymbol{z}_\text{occ}^\text{seq} \in \mathbb{R}^{T \times C \times h \times w}$ into an occupancy sequence $\mathbf{O}^\text{seq} \in \mathbb{R}^{T \times H \times W \times D}$. It is built using 3D convolutional layers and 3D axial attention layers to capture temporal features. Similar to~\cite{zheng2023occworld}, we use cross-entropy loss $\mathcal{L}_\text{CE}$, Lovasz-softmax loss $\mathcal{L}_\text{LS}$, and KL divergence loss $\mathcal{L}_\text{KL}$ to train the VAE. The total loss for the occupancy VAE is:
\begin{equation}
\begin{aligned}
   \mathcal{L}_\text{occ}^\text{vae} &= \mathcal{L}_\text{CE}+ \lambda_1 \mathcal{L}_\text{LS} +  \lambda_2 \mathcal{L}_\text{KL}, 
\end{aligned}
\end{equation}
where $\lambda_1$ and $\lambda_2$ are the respective loss weights. Additional details about the training setup for the occupancy VAE can be found in the supplementary materials.

\noindent\textbf{Latent Occupancy DiT.}
The Latent Occupancy DiT learns to generate occupancy latent sequence
from noise volumes with the condition of the BEV layout $\mathbf{B}$.
Specifically, the BEV layouts are first concatenated with the noise volumes, which are further patchified before being fed into the Occupancy DiT. This explicit alignment strategy helps the model efficiently learn spatial relationships, enabling more precise control over the generated sequences.
The Occupancy DiT aggregates spatial-temporal information through a series of stacked spatial and temporal transformer blocks \cite{ma2024latte}.
The loss function of Occupancy DiT is defined following ~\cite{peebles2023scalable}:
\vspace{-3pt}
\begin{equation}
\mathcal{L}_\text{occ}^\text{dit}=
\mathbb{E}\left[\sum_{i=1}^T\left \| \boldsymbol{f}_\text{dit}\left(\boldsymbol{z}_\text{occ}^{i}, \mathbf{B}^{i}\right)- \boldsymbol{\epsilon}_\text{n}^i \right\|^2    
\right],
\end{equation}
where $\boldsymbol{f}_\text{dit}\left(\boldsymbol{z}_\text{occ}^{i}, \mathbf{B}^{i} \right)$ represents the model output, while $\boldsymbol{z}_\text{occ}^{i}$ represents the input noisy latent of $i^{th}$ frame. $\boldsymbol{\epsilon}_\text{n}^i$ is the random noise sampled from $\mathcal{N}(\mathbf{0}, \boldsymbol{I})$.
By incorporating temporal information, our occupancy diffusion model enables the generation of long-term consistent occupancy sequences (see Fig.~\ref{teaser_fig1_b} (a)). More architectural details about the Occupancy VAE and DiT can be found in the supplementary.

\subsection{Video: Occupancy as Conditional Guidance}\label{sec_video}

The video generation model is initialized with the pre-trained latent generation model of Stable Video Diffusion (SVD)~\cite{blattmann2023stable}, which is composed of a 3D video VAE and a video diffusion UNet. 
As depicted in Fig.~\ref{fig_overall}, the video diffusion UNet takes occupancy-based rendering maps and text prompts as conditions to generate multi-view driving videos.

\noindent\textbf{Gaussian-based Joint Rendering.}
The input semantic occupancy grids are jointly rendered into multi-view semantic and depth maps with forward Gaussian Splatting~\cite{kerbl20233d,zhou2024hugs}. 
The rendered maps bridge the representation gaps between the occupancy grids and the multi-view video, providing fine-grained semantic and geometric guidance to facilitate high-quality and consistent video generation.
Rather than relying on the resource-intensive spatial-temporal attention mechanisms used in previous works~\cite{gao2023magicdrive,wu2023tune} for cross-view consistency, we retain the vanilla cross-attention from SVD~\cite{blattmann2023stable} and ensure cross-view consistency through occupancy-based multi-view conditional guidance. Additional experimental evaluations are provided in Fig.\ref{fig_video_ablation} and Tab.\ref{table_ar_video}. 

\begin{figure}[!t]
    \centering
        \vspace{-6pt}
    \includegraphics[width=1\linewidth]{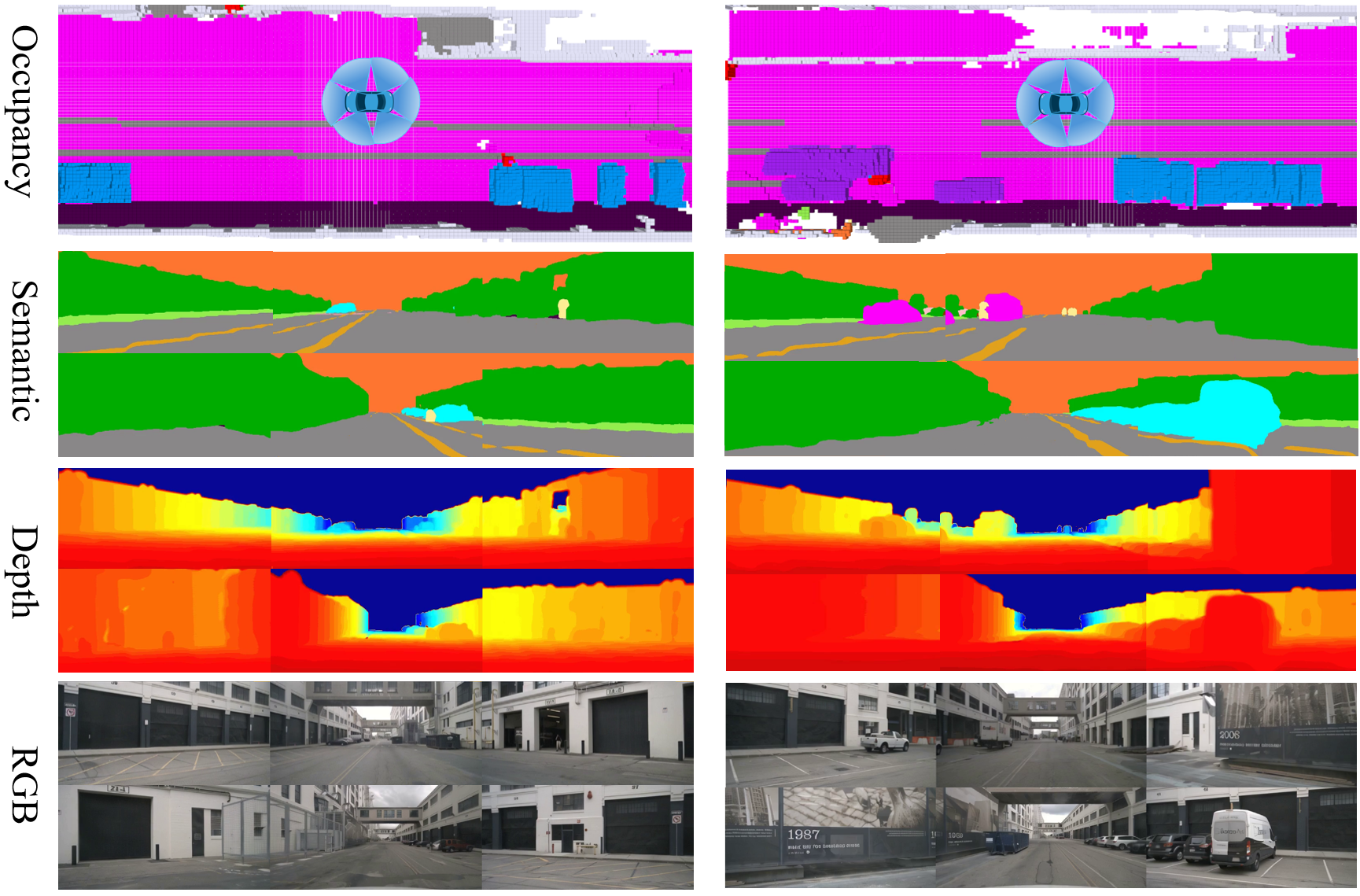}
    \vspace{-16pt}
    \caption{\textbf{Visualization} of the Gaussian-based joint rendering. }
    \label{fig_joint_rend}
    \vspace{-18pt}
\end{figure}

Specifically, given input semantic occupancy with the shape of $\mathbb{R}^{H \times W \times D}$, we first convert it into a set of 3D Gaussian primitives $\mathcal{G} = \{G_i\}_{i=1}^{N}$ based on the center and semantic label of each occupancy grid.
In this way, each Gaussian primitive contains attributes of position ${\mu}$, semantic label ${s}$, opacity status $\alpha$, and the covariance ${\Sigma}$ calculated with the default rotation and voxel size scaling. 
Then, the depth map $\mathbf{D}$ and the semantic map $\mathbf{S}$ are rendered via tile-based rasterization~\cite{kerbl20233d} similar to color rendering:
\vspace{-5pt}
\begin{equation} 
\mathbf{D}=\sum_{i \in {N}} {d}_i \alpha_i^{\prime} \prod_{j=1}^{i-1}\left(1-\alpha_j^{\prime}\right),
\end{equation}
\vspace{-10pt}
\begin{equation} 
\mathbf{S}=\texttt{argmax} ( \sum_{i \in {N}} \texttt{onehot}({s}_i) \alpha_i^{\prime} \prod_{j=1}^{i-1}\left(1-\alpha_j^{\prime}\right)),
\end{equation}
where $d_i$ denotes the depth value and $\alpha^{\prime}$ is determined by the projected 2D Gaussian and
the 3D opacity $\alpha$ as~\cite{kerbl20233d}.
The visualization results of the rendered semantic and depth maps are shown in Fig.~\ref{fig_joint_rend}. Note that the road lines from the BEV layouts are projected onto the semantic occupancy, integrating the corresponding semantic information. These maps are fed into an encoder branch with residual connections and zero convolutions, similar to ControlNet~\cite{zhang2023adding}, to leverage the pre-trained capabilities of the video diffusion UNet while preserving its inherent generative power.
 
\noindent\textbf{Geometric-aware Noise Prior.}
To further enhance video generation quality, we introduce a geometry-aware noise prior strategy during the sampling process. It injects a dense appearance prior, similar to previous works~\cite{wang2024microcinema,gao2024scp}, while also incorporating explicit geometric awareness through the rendered depth map $\mathbf{D}$ to model regional correlations.

\vspace{-5pt}
\begin{equation}\label{noise_prior} 
\boldsymbol{\epsilon}_\text{vid}^i=\lambda \boldsymbol{z}_{c}+ \boldsymbol{\epsilon}_\text{n}^i,
\end{equation} 
where $\boldsymbol{\epsilon}_\text{vid}^i$ represents the noise input of $i^{th}$ video frame. 
$\boldsymbol{z}_{c}$ denotes the latent feature corresponding to the conditional video frame, obtained by encoding the conditional video frame using the 3D video VAE encoder.
$\boldsymbol{\epsilon}_\text{n}^i$ is random noise sampled from $\mathcal{N}(\mathbf{0}, \boldsymbol{I})$.
$\lambda$ is the balancing coefficient. 

Nevertheless, in real-world scenarios, many regions within dynamic videos exhibit significant variations across multiple frames. The simple strategy described earlier does not account for correspondence modeling in these highly dynamic regions. To address this, we leverage the rendered depth map $\mathbf{D}$ to warp the appearance prior from the reference image to other images using depth-based reprojection~\cite{cai2023riav,watson2021temporal}, enabling explicit geometric awareness. With this approach, we revisit Eq.~\ref{noise_prior} and optimize it as follows:
\begin{equation}\label{noise_prior_2} 
\boldsymbol{\epsilon}_\text{vid}^i=\lambda (\texttt{Warp} (\boldsymbol{z}_{c}, \mathbf{D}^{i},\mathbf{K}, \left[\mathbf{R}_{0, i} \mid \mathbf{t}_{0, i}\right]  )) + \boldsymbol{\epsilon}_\text{n}^i,
\end{equation} 
where the depth-based $\texttt{Warp}$ process~\cite{cai2023riav,watson2021temporal} projects the pixel $p_i={(u,v,1)}^T$ in the $i^{th}$ latent feature $z_i$ to its counterpart ${p}_0$ in the conditional latent feature $z_c$ as:
\vspace{-5pt}
\begin{equation} 
{p}_0 = \mathbf{K} \cdot (\mathbf{R}_{0, i} \cdot ({\mathbf{K}}^{-1} \cdot p_i \cdot \mathbf{D}^{i}(u,v)) + \mathbf{t}_{0, i} ). 
\vspace{-5pt}
\end{equation} 
where 
$\mathbf{K}$ is the camera intrinsics of the latent features, scaled from the original sensor parameters.
$\left[\mathbf{R}_{0, i} \mid \mathbf{t}_{0, i}\right]$ is the transform matrix from target latent $z_i$ to conditional latent $z_c$.
$\mathbf{D}^{i}$ is the rendered depth map of $i^{th}$ video frame. 
More details of the evaluation results can be found in Tab.~\ref{table_ar_video} and the supplementary.

\noindent\textbf{Video Training Loss.}
We define the video training loss similar to previous works~\cite{blattmann2023stable,gao2024vista}, which is formulated as:
\vspace{-5pt}
\begin{equation}
\mathcal{L}_\text{vid}=\mathbb{E}\left[\sum_{i=1}^T\left(1-m^i\right) \cdot\left\|\boldsymbol{f}_\text{vid}\left( \boldsymbol{z}_\text{vid}^i, t, \boldsymbol{z}_{c}, \mathbf{D}^{i}, \mathbf{S}^{i} \right)-\boldsymbol{z}_0^i\right\|^2\right],
\end{equation}
where 
$\boldsymbol{f}_\text{vid}\left(\boldsymbol{z}_\text{vid}^i, t, \boldsymbol{z}_{c}, \mathbf{D}^{i}, \mathbf{S}^{i}\right)$ represents the output of the video generation model.
$\mathbf{D}^{i}$ and $\mathbf{S}^{i}$ denote the rendered depth map and semantic map of $i^{th}$ video frame, respectively. $t$ represents the input text prompt.
$\boldsymbol{z}_0^i, \boldsymbol{z}_\text{vid}^i$ are the ground truth and input noisy latent, respectively.
$\boldsymbol{z}_c$ is the conditional reference frame leveraged following SVD~\cite{blattmann2023stable}.
$m$ is the one-hot mask used to select the condition frames.
Note that we randomly select $\boldsymbol{z}_c$ to diminish the model's reliance on any specific conditioning frame.

\subsection{LiDAR: Occupancy-based Sparse Modeling }\label{sec_lidar}

{As illustrated in Fig.~\ref{fig_overall}, for LiDAR generation, the input occupancy is first encoded into sparse voxel features with a Sparse UNet~\cite{shi2020points}, which are then used to produce LiDAR points via sparse sampling guided by occupancy priors.}

\noindent\textbf{Prior Guided Sparse Modeling.} 
Given the inherent sparsity and detailed geometry of semantic occupancy, we propose a prior-guided sparse modeling approach to enhance computational efficiency (see the last column of Tab.\ref{tab_lidar}), by avoiding unnecessary computations on unoccupied voxels. The input semantic occupancy grids are first processed using a Sparse UNet\cite{shi2020points} to aggregate contextual features. Next, we perform uniform sampling along the LiDAR rays, denoted as $\mathbf{r}$, to generate a sequence of points, represented as $s$.

\begin{figure}[!t]
    \centering
    \vspace{-7pt}
    \includegraphics[width=1.0\linewidth]{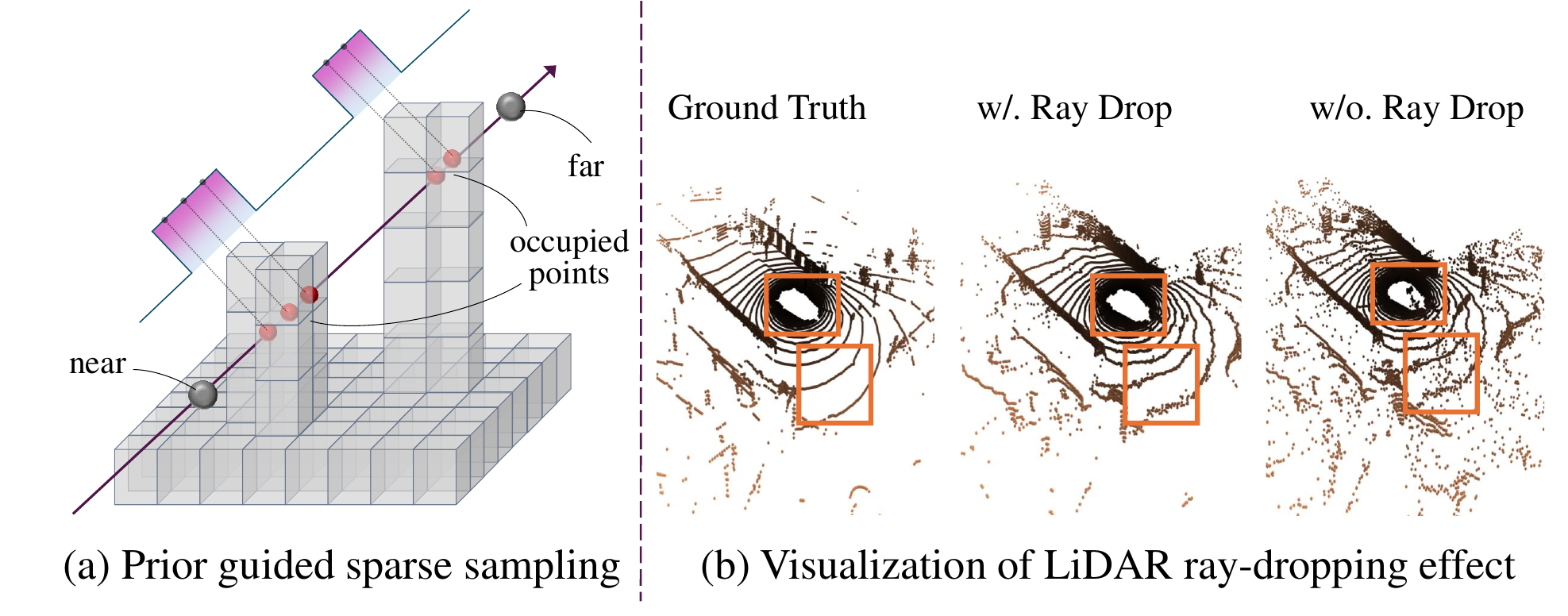}
    \vspace{-17pt}
    \caption{(a) \textbf{Sparse sampling} with occupancy-based prior guidance. (b) \textbf{Visualization} of the effect on LiDAR ray-dropping head.}
    \label{fig_lidar_sparse_sample}
    \vspace{-15pt}
\end{figure}

As shown in Fig.~\ref{fig_lidar_sparse_sample} (a), to facilitate prior-guided sparse sampling, we assign a probability of 1 to points within occupied voxels and 0 to all other points, thus defining a probability distribution function (PDF). Subsequently, we resample $n$ points 
$\{\mathbf{r}_i={o}+s_i{v} \ (i=1,...,n)\}$ based on the PDF.
Here ${o}$ is the ray origin and ${v}$ is the normalized ray direction.

\noindent\textbf{LiDAR Head and Training Loss.} 
Following the ray-based volume rendering techniques from previous works~\cite{unisim, Barron_2023_ICCV, wang2021neus}, the features of each resampled point are processed through a multi-layer perceptron (MLP) to predict the signed distance function (SDF) $f(s)$ and to compute the associated weights $w(s)$. These predictions and weights are then used to estimate the depth of the ray via volume rendering:
\begin{equation}
    \beta_i = \max(\frac{\Phi_s(f(\mathbf{r}(s_i)))-\Phi_s(f(\mathbf{r}(s_{i+1})))}{\Phi_s(f(\mathbf{r}(s_i)))}, 0),
\end{equation}
\vspace{-10pt}
\begin{equation}
    w(s_i)=\prod_{j=1}^{i-1}(1-\beta_j)\beta_i, \quad h = \sum_{i=1}^n w(s_i)s_i,
\end{equation}
where $\Phi_s(x)=(1+e^{-sx})^{-1}$, $h$ is the rendered depth value.  
To more accurately simulate the realistic LiDAR imaging process, we incorporate an additional reflection intensity head and a ray-dropping head. The reflection intensity head predicts the intensity of the object's reflection of the LiDAR laser beam along each ray. This prediction involves a weighted sum of the point features along the ray according to $w(s)$, followed by an MLP for estimation. 
The ray-dropping model estimates the probability of a ray not being captured by the LiDAR due to the failure in detecting reflected light, which is implemented with the same structure as the reflection intensity head. As shown in Fig.~\ref{fig_lidar_sparse_sample} (b), the ray-dropping head effectively eliminates noise points in the predictions.
The training loss for LiDAR generation is composed of depth loss $\mathcal{L}_\text{depth}$, intensity loss $\mathcal{L}_\text{inten}$ and ray-dropping loss ${L}_\text{drop}$:

\vspace{-10pt}
\begin{equation}
\begin{aligned}
\mathcal{L}_\text{lid}
&= \mathcal{L}_\text{depth}+ \lambda_1\mathcal{L}_\text{inten}+\lambda_2\mathcal{L}_\text{drop} 
,
\end{aligned}
\end{equation}
where 
$\lambda_1$, $\lambda_2$ are balancing coefficients. More details including the training setup are provided in the supplementary.

\begin{figure*}[!t]
 \vspace{-39pt}
    \centering
    \includegraphics[width=0.92\linewidth]{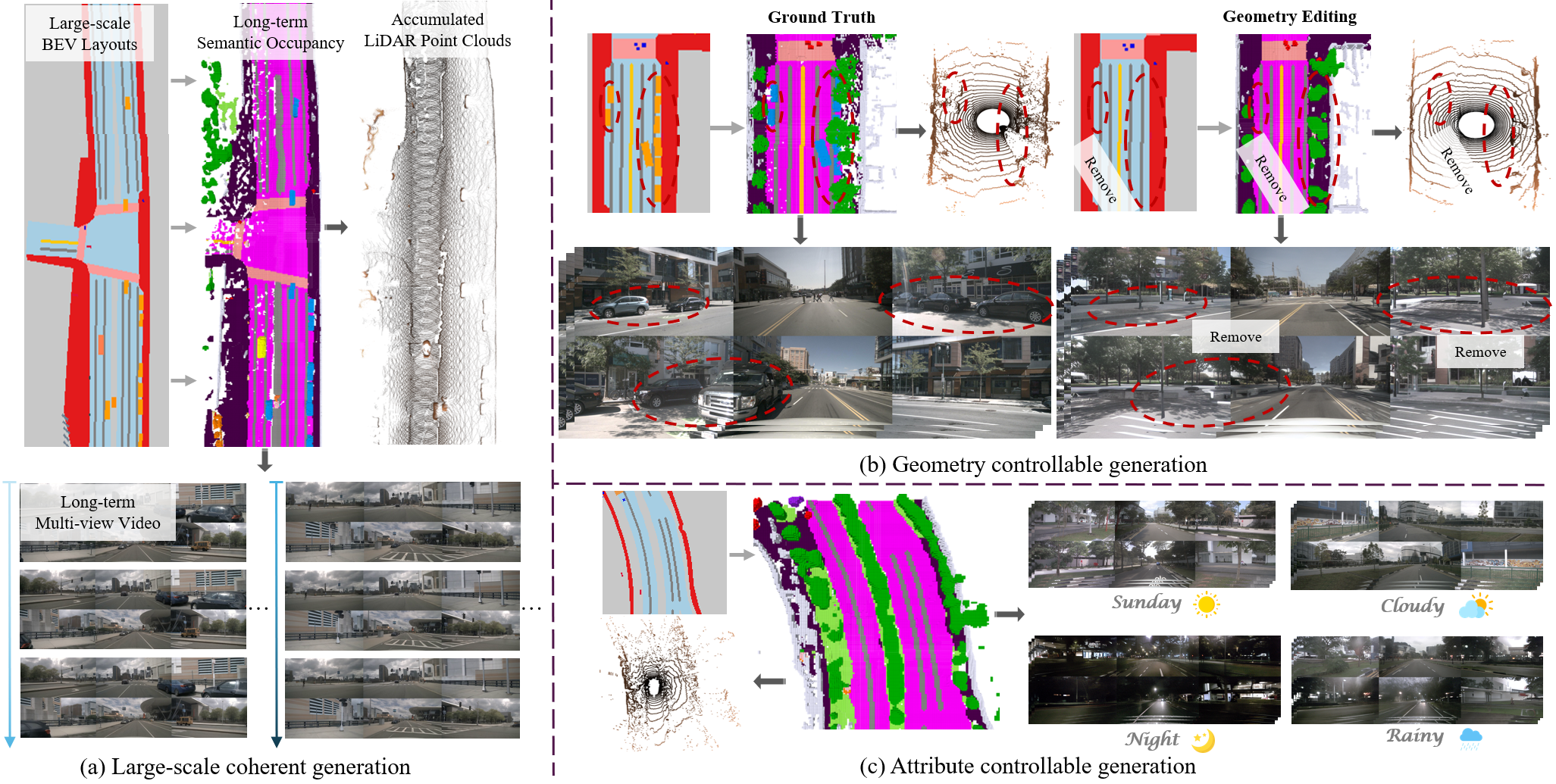}
     \vspace{-10pt}
    \caption{\textbf{Versatile generation ability of UniScene.} (a) Large-scale coherent generation of semantic occupancy, LiDAR point clouds, and multi-view videos. 
    (b) Controllable generation of geometry-edited occupancy, video, and LiDAR by simply editing the input BEV layouts to convey user commands.
    (c) Controllable generation of attribute-diverse videos by changing the input text prompts.
    }
    \label{teaser_fig1_b}
    \vspace{-15pt}
\end{figure*}

\vspace{-2pt}
\section{Experiments}
Our experiments are conducted on the challenging NuScenes benchmark~\cite{caesar2020nuscenes}. We interpolate the semantic occupancy labels from the 2Hz key-frame annotations of the NuScenes-Occupancy dataset~\cite{wang2023openoccupancy} to a higher frame rate of 12Hz, similar to previous studies~\cite{wang2022asap,zhao2024drivedreamer,gao2023magicdrive}. For additional details, including dataset specifics, evaluation metrics, comparison baselines, model configurations, and further visualizations, please refer to the supplementary materials.

\subsection{Main Results}

\textbf{Versatile Generation Ability.} As shown in Fig.~\ref{teaser_fig1_b}, the proposed UniScene facilitates versatile and controllable generation.
For large-scale coherent generation, UniScene first produces long-term semantic occupancy from the given BEV layouts, which subsequently guides the generation of LiDAR point clouds and multi-view videos (Fig.~\ref{teaser_fig1_b}(a)).
Regarding controllable geometry editing, users can easily edit the BEV layouts, such as by removing cars (Fig.~\ref{teaser_fig1_b}(b)), and the generation results can change accordingly.
Furthermore, as depicted in Fig.~\ref{teaser_fig1_b}(c), for attribute customization, UniScene allows for varied weather and lighting conditions in the generated videos through user-specified text prompts (\eg, specifying weather or time of day).

\begin{table}[!t]
\vspace{-0pt}
\begin{center}
\scriptsize
\vspace{-0pt}
\renewcommand\tabcolsep{7.6pt}
	\centering
   \resizebox{0.99\linewidth}{!}{
\begin{tabular}{l|c|c|c}
\toprule Method  & 
\begin{tabular}[c]{@{}c@{}}Compression\\Ratio\end{tabular} $\uparrow$ & mIoU $\uparrow$ & IoU $\uparrow$ \\ \midrule
OccLLama (VQVAE)~\cite{wei2024occllama} & 8 & \underline{75.2}& 63.8\\
OccWorld (VQVAE)~\cite{zheng2023occworld}  & 16 & 65.7 & 62.2 \\
OccSora (VQVAE)~\cite{wang2024occsora} & \textbf{512} & 27.4& 37.0\\
\midrule
\rowcolor{gray!10} Ours (VAE)  &  \underline{32}  & \textbf{92.1} &  \textbf{87.0} \\
\rowcolor{gray!10} Ours (VAE) & \textbf{512} & 72.9 & \underline{64.1}\\
\bottomrule
\end{tabular}
 }
 \vspace{-8pt}
 \caption{\textbf{Occupancy Reconstruction.} 
 The compression ratio is calculated following the methodology outlined in OccWorld~\cite{zheng2023occworld}. The top two performers are highlighted in \textbf{bold} and \underline{underlined}.
 }
\vspace{-18pt}
\label{tab_occ_rec}
\end{center}
\end{table}

\begin{table}[!t]
\vspace{-0pt}
\begin{center}
\scriptsize
\vspace{-0pt}
\renewcommand\tabcolsep{9.6pt}
	\centering
   \resizebox{0.99\linewidth}{!}{
\begin{tabular}{l|cccc}
    \toprule
    Method & CFG   & mIoU $\uparrow$ &  F3D $\downarrow$  & MMD $\downarrow$ \\
    \midrule
    \multirow{2}[1]{*}{Ours-Gen.} & 4     & \underline{20.51} & 205.78 & 11.60 \\
          & 1     & 19.44 & 158.55 & 10.60\\
    \midrule
    OccWorld \cite{zheng2023occworld}& -     & {17.13} & \underline{145.65} & \underline{9.89} \\
    \rowcolor{gray!10}Ours-Fore. & -     & \textbf{31.76} & \textbf{43.13} & \textbf{2.86} \\
    \bottomrule
    \end{tabular}%
 }
 \vspace{-8pt}
 \caption{\textbf{Occupancy Generation and Forecasting. } 
`Ours-Gen.' and `Ours-Fore.' denote our Generation model and Forecasting model, respectively. `CFG' refers to the Classifier-Free Guidance.
}
\vspace{-25pt}
\label{occ_gen_fidelity}
\end{center}
\end{table}

\noindent\textbf{Occupancy Reconstruction, Generation and Forecasting.}\label{exp_occ}
The evaluation of occupancy is on the NuScenes-Occupancy key-frame (2Hz) validation set.
Since precisely reconstructing the occupancy is vital for occupancy generation, we first compare our Occupancy VAE with the existing methods on reconstruction accuracy in Tab.~\ref{tab_occ_rec}. 
Compared to the discrete compression with VQVAE in previous works~\cite{wei2024occllama,zheng2023occworld,wang2024occsora}, our continuous compression with VAE achieves remarkable reconstruction performance even under the high compression ratio of 512, 
surpassing OccWorld~\cite{zheng2023occworld} by 10.96\% in mIoU. 
Note that the compression ratio of 512 is employed as the default setting in our model to ensure efficiency.

\begin{figure}[!t]
    \centering
    \vspace{-5pt}
    \includegraphics[width=1.0\linewidth]{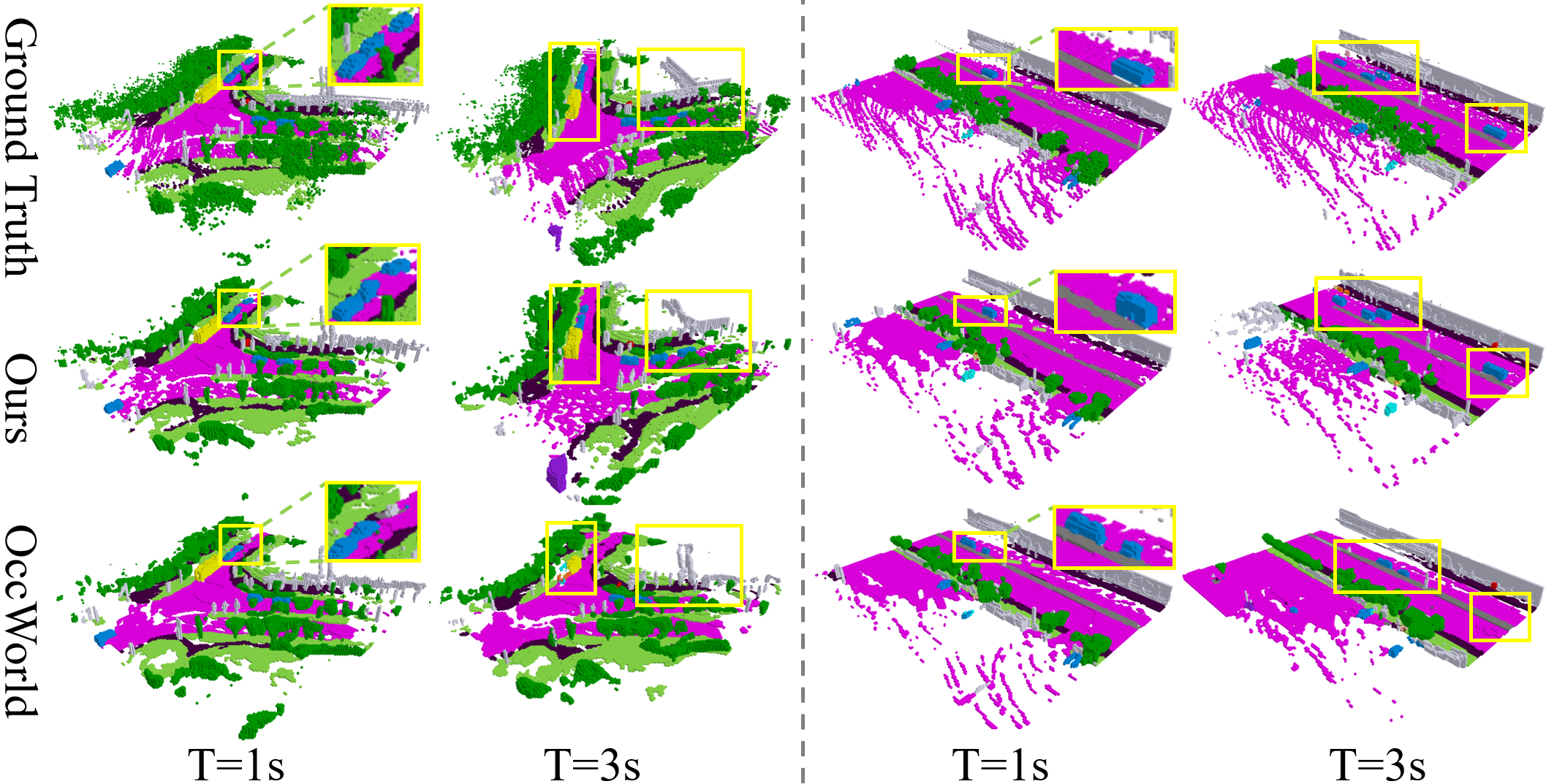}
    \vspace{-17pt}
    \caption{\textbf{Qualitative evaluation} for occupancy forecasting. 
    Our method can compellingly handle sharp steering maneuvers and dynamic objects with temporal consistency.
    }
    \label{occ-wm-compare}
    \vspace{-15pt}
\end{figure}

To further evaluate the effectiveness of our occupancy generation model, we implement a forecasting variant of our model as previous forecasting works~\cite{zheng2023occworld} for comparison. Please refer to the supplementary for forecasting adaption details.
The quantitative evaluation for occupancy generation (`Ours-Gen.') and forecasting (`Ours-Fore.') is shown in Tab.~\ref{occ_gen_fidelity}.
Even with fewer reference occupancy frames (`Ours-Fore.': 2, OccWorld: 5), `Ours-Fore' surpasses OccWorld by 70.39\% in F3D and 71.08\% in MMD, respectively.
The qualitative results of occupancy forecasting are shown in Fig.~\ref{occ-wm-compare}.
Our method can handle dynamic objects and sharp steering maneuvers compellingly in the generating process.

\begin{table}[!t]
\vspace{-0pt}
\begin{center}
\scriptsize
\vspace{-0pt}
\renewcommand\tabcolsep{7.5pt}
\centering
\resizebox{0.99\linewidth}{!}{
\begin{tabular}{l|cc|cc}
\toprule Method & Multi-view & Video & FID $\downarrow$ & FVD $\downarrow$ \\ \midrule
BEVGen~\cite{swerdlow2024streetview} & \textcolor{ForestGreen}{\usym{2713}} & \textcolor{red}{\usym{2717}} & 25.54 & - \\
BEVControl~\cite{yang2023bevcontrol} & \textcolor{ForestGreen}{\usym{2713}} & \textcolor{red}{\usym{2717}} & 24.85 & - \\
 DriveGAN~\cite{kim2021drivegan} & \textcolor{red}{\usym{2717}} & \textcolor{ForestGreen}{\usym{2713}} & 73.40 & 502.30 \\
DriveDreamer~\cite{zhao2024drivedreamer} & \textcolor{red}{\usym{2717}} & \textcolor{ForestGreen}{\usym{2713}} & 52.60 & 452.00 \\
Vista~\cite{gao2024vista} & \textcolor{red}{\usym{2717}} & \textcolor{ForestGreen}{\usym{2713}}& 6.90 & 89.40 \\ \midrule

WoVoGen~\cite{lu2023wovogen} & \textcolor{ForestGreen}{\usym{2713}} & \textcolor{ForestGreen}{\usym{2713}} & 27.60 & 417.70 \\
Panacea~\cite{wen2023panacea} & \textcolor{ForestGreen}{\usym{2713}} & \textcolor{ForestGreen}{\usym{2713}} & 16.96 & 139.00\\
MagicDrive~\cite{gao2023magicdrive} & \textcolor{ForestGreen}{\usym{2713}} & \textcolor{ForestGreen}{\usym{2713}} & 16.20 & - \\  
Drive-WM~\cite{wang2023driving} & \textcolor{ForestGreen}{\usym{2713}} & \textcolor{ForestGreen}{\usym{2713}} &  15.80  &  122.70  \\
Vista $^*$~\cite{gao2024vista} & \textcolor{ForestGreen}{\usym{2713}} & \textcolor{ForestGreen}{\usym{2713}}& 13.97 & 112.65  \\ \midrule
\rowcolor{gray!10}Ours (Gen Occ) & \textcolor{ForestGreen}{\usym{2713}}&  \textcolor{ForestGreen}{\usym{2713}}& \underline{6.45} & \underline{71.94} \\ 
\rowcolor{gray!10}Ours (GT Occ) & \textcolor{ForestGreen}{\usym{2713}}&  \textcolor{ForestGreen}{\usym{2713}}& \textbf{6.12} & \textbf{70.52} \\
\bottomrule
\end{tabular}
 }
 \vspace{-8pt}
\caption{\textbf{Video Generation.} 
We implement the multi-view variant of Vista$^*$~\cite{gao2024vista} with spatial-temporal attention~\cite{wu2023tune,gao2023magicdrive}.
}
\vspace{-20pt}
\label{tab_video}
\end{center}
\end{table}

\begin{table}[!t]
\vspace{-0pt}
\begin{center}
\scriptsize
\vspace{-0pt}
\renewcommand\tabcolsep{9.9pt}
	\centering
   \resizebox{1.00\linewidth}{!}{
\begin{tabular}{l|ccc}
\toprule Method  & MMD ($10^{-4}$)$\downarrow$ & JSD $\downarrow$ & Time (s)$\downarrow$  \\ \midrule
LiDARDM~\cite{zyrianov2024lidardm} & 3.51 & 0.118 & 45.12 \\
Open3D~\cite{zhou2018open3d}  & 8.15& 0.149 & 2.39   \\ 
\midrule
\rowcolor{gray!10}Ours (Gen Occ)    & \underline{2.40}  &  \underline{0.108} & \underline{0.47}  \\
\rowcolor{gray!10}Ours (GT Occ)    & \textbf{1.53} & \textbf{0.072} & \textbf{0.25}   \\
\bottomrule
\end{tabular}
 }
 \vspace{-8pt}
 \caption{\textbf{LiDAR Generation.} 
 We include the semantic occupancy generation time for a fair comparison.
}
\vspace{-25pt}
\label{tab_lidar}
\end{center}
\end{table}

\noindent\textbf{Video Generation Results.}
The quantitative comparison of video generation is illustrated in Tab.~\ref{tab_video}. 
The `Gen Occ' and `GT Occ' represent the occupancy conditions generated from our framework and ground truth, respectively.
Our method supports multi-view video generation and outperforms all the other methods, achieving 71.94 FVD with generated occupancy and 70.52 FVD with ground truth occupancy, respectively.
Note that the competitive method of Vista~\cite{gao2024vista} only supports single-view video generation, here we implement the multi-view variant with spatial-temporal attention following~\cite{wu2023tune,gao2023magicdrive} for a fair comparison.

\begin{figure}[!t]
    \centering
    \vspace{-5pt}
    \includegraphics[width=1.0\linewidth]{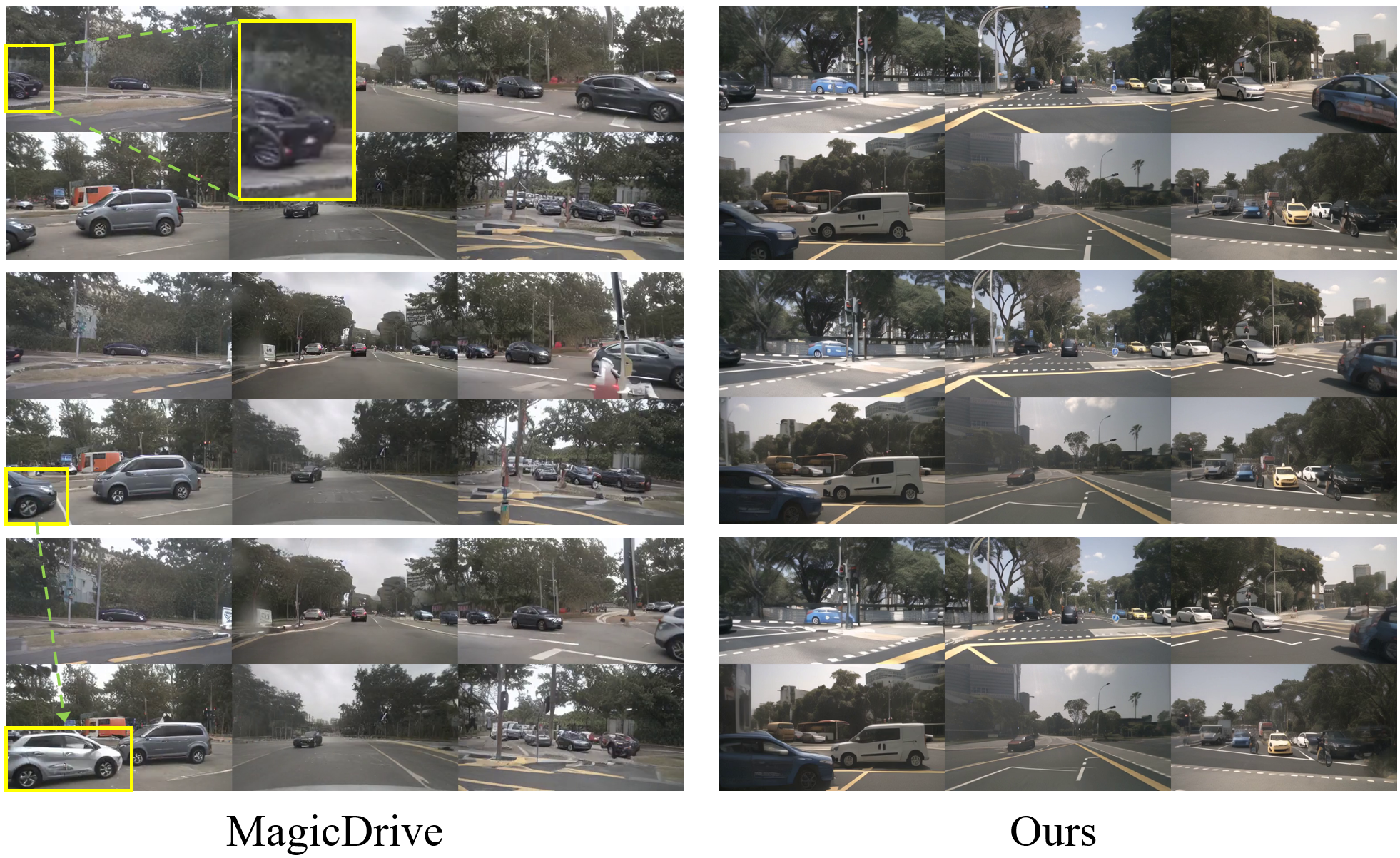}
    \vspace{-18pt}
    \caption{\textbf{Qualitative evaluation} for video generation. Our method excels in object structure quality and cross-view consistency.}
    \label{video_compare}
    \vspace{-6pt}
\end{figure}

As shown in Fig.~\ref{video_compare}, we compare our video generation results with the video generation model of MagicDrive~\cite{gao2023magicdrive}. 
Our approach demonstrates an obvious improvement in video generation quality, particularly in the structure quality of objects and cross-view consistency. The notable enhancement is attributed to the conditional guidance derived from occupancy-based semantic and geometric rendering maps.

\begin{figure}[!t]
    \centering
    \includegraphics[width=1.0\linewidth]{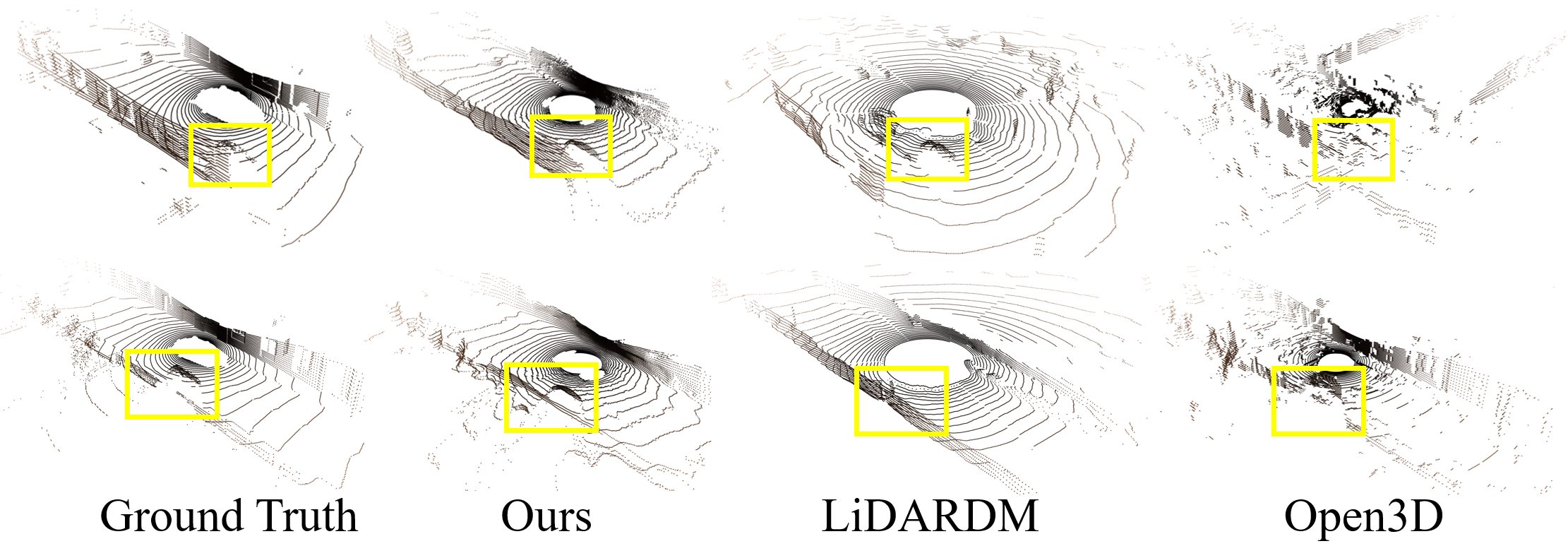}
    \vspace{-10pt}
    \caption{\textbf{Qualitative evaluation} for LiDAR generation. Our method generates precious scene layouts and clear object details.}
    \label{lidar_compare}
    \vspace{-10pt}
\end{figure}

\noindent\textbf{LiDAR Generation Results.}
We compare our LiDAR generation model with Open3D~\cite{zhou2018open3d} and LiDARDM~\cite{zyrianov2024lidardm} on the NuScenes validation set.
For Open3D, we utilize the library's hard ray-casting function to convert the ground truth occupancy grids into corresponding LiDAR point clouds. LiDARDM is implemented using the official repository and trained with the same setting as our model on NuScenes.
The `Gen Occ' and `GT Occ' are the same as the video generation evaluation.
As shown in Tab.~\ref{tab_lidar}, our method achieves the best generation performance, surpassing LiDARDM by 31.62\% in MMD.
Moreover, our approach demonstrates significant advantages in inference speed over other methods, which stem from our efficient sparse modeling scheme.
The qualitative results are shown in Fig.~\ref{lidar_compare}.
Compared to LiDARM and Open3D, our method shows significant superiority in generating precious scene layouts and clear object details.

\noindent\textbf{Downstream Task Evaluation.} 
UniScene can produce multi-modal augmented data with corresponding annotations, thereby enhancing training for downstream perception tasks. We augment an equal quantity of images and LiDAR point clouds as in the original NuScenes dataset, ensuring consistent training iterations and batch sizes for fair comparisons as previous works~\cite{gao2023magicdrive,swerdlow2024streetview}. Unlike existing works~\cite{gao2023magicdrive,swerdlow2024streetview} that only produce augmented images, we generate different data forms to enable the training of comprehensive downstream multi-modal perception models. 
For the Occupancy Prediction task in Tab.~\ref{tab_down1}, UniScene significantly improves the baseline of CONet~\cite{wang2023openoccupancy} in different modality settings (\eg, Camera (C), LiDAR (L), Camera+LiDAR (C\&L) ), which stems from the high fidelity generation with precious occupancy-based conditional guidance of fine-grained geometry and semantics. 
For the BEV Segmentation and Object Detection tasks in Tab.~\ref{tab_down2}, we employ CVT~\cite{zhou2022cross}) and BEVFusion~\cite{liu2023bevfusion} as baselines following~\cite{gao2023magicdrive}. Our method significantly surpasses other methods for data augmentation, highlighting the superior generation quality.

\subsection{Ablation Studies}\label{ablation}

\noindent\textbf{Effect of Designs in Occupancy Generation Model.}
We conduct extensive ablation studies to validate the effectiveness of the occupancy generation model components, as shown in Tab.~\ref{table_ab_occ_1}. 
Incorporating temporal information in the occupancy VAE decoder with 3D axial attention boosts occupancy sequence generation fidelity, particularly reducing the MMD metric by 38.01\%. 
The temporal attention layer in Occupancy DiT effectively improves the generation results, boosting the F3D metric by 10.29\%. 
The spatial attention layer in Occupancy DiT brings significant enhancement in generation fidelity, reducing the F3D metric by 39.26\%.

\begin{table}[!t]
\begin{center}
\scriptsize
\vspace{-0pt}
\renewcommand\tabcolsep{12.0pt}
	\centering
   \resizebox{1.00\linewidth}{!}{
	\begin{tabular}{l| c| c c }
 
		\toprule
		Method & \makecell[c]{Input} & \makecell[c]{IoU} $\uparrow$ & \makecell[c]{mIoU} $\uparrow$
		  \\
		\midrule
MonoScene~\cite{cao2022monoscene} & $C$  & 18.4 & 6.9 \\
TPVFormer~\cite{huang2023tri}   & $C$ & 15.3 & 7.8  \\ 
Baseline-C~\cite{wang2023openoccupancy} & $C$ & 20.1 & 12.8 \\
w/. Vista$^*$~\cite{gao2024vista} & $C$ & 20.8$\mathrm{\tiny{\hi{+0.7}}}$ & 13.1$\mathrm{\tiny{\hi{+0.3}}}$  \\  
w/. MigicDrive~\cite{gao2023magicdrive} & $C$ & 21.8$\mathrm{\tiny{\hi{+1.7}}}$ & 13.9$\mathrm{\tiny{\hi{+1.1}}}$   \\ 
\rowcolor{gray!10}w/. Ours-C & $C$ &  \textbf{28.6}$\mathrm{\tiny{\hi{+8.5}}}$ & \textbf{16.5}$\mathrm{\tiny{\hi{+3.7}}}$ \\ \midrule

 LMSCNet~\cite{roldao2020lmscnet}   & $L$ &27.3  & 11.5  \\ 
 JS3C-Net~\cite{yan2021sparse}   & $L$ &30.2  & 12.5  \\
Baseline-L~\cite{wang2023openoccupancy} & $L$ & 30.9 & 15.8  \\
\rowcolor{gray!10}w/. Ours-L & $L$ & \textbf{33.1}$\mathrm{\tiny{\hi{+2.2}}}$ & \textbf{19.3}$\mathrm{\tiny{\hi{+3.5}}}$    \\ \midrule

3DSketch~\cite{chen20203d}   & $C \& L^D$ &25.6 & 10.7   \\  
AICNet~\cite{li2020anisotropic} & $C \& L^D$ & 23.8 & 10.6  \\ 
Baseline-M~\cite{wang2023openoccupancy} & $C \& L$& 29.5 & 20.1 \\              
\rowcolor{gray!10}w/. Ours-M & $C \& L$ & \textbf{35.4}$\mathrm{\tiny{\hi{+5.9}}}$ & \textbf{23.9}$\mathrm{\tiny{\hi{+3.8}}}$  
\\  \bottomrule
\end{tabular}
}
\vspace{-8pt}
\caption{
\textbf{Comparison about training support for semantic occupancy prediction} (Baseline as CONet~\cite{wang2023openoccupancy}). 
The ``$C$'', ``$L$'', and ``$L^D$'' denote the camera, LiDAR, and depth projected from LiDAR. 
}
\vspace{-18pt}
\label{tab_down1}
\end{center}
\end{table}

\begin{table}[!t]
\vspace{-0pt}
\begin{center}
\scriptsize
\vspace{-0pt}
\renewcommand\tabcolsep{1.0pt}
\resizebox{1.00\linewidth}{!}{%
\begin{tabular}{l|c|cc|cc}
\toprule \multicolumn{1}{l|}{\multirow{2}{*}{Method}} & \multicolumn{1}{c|}{\multirow{2}{*}{Input}}  & \multicolumn{2}{c|}{BEV Segmentation} & \multicolumn{2}{c}{3D Object Detection} \\  
 \multicolumn{1}{c|}{} & & Road mIoU $\uparrow$ & Vehicle mIoU $\uparrow$ & mAP $\uparrow$ & NDS $\uparrow$ \\  \midrule  
Baseline~\cite{zhou2022cross,liu2023bevfusion} & $C$ & 74.30 & 36.00  & 32.88 & 37.81   \\ 
w/. BEVGen~\cite{swerdlow2024streetview} & $C$  & 71.90$\mathrm{\tiny{\hi{-2.40}}}$ & 36.60$\mathrm{\tiny{\hi{+0.60}}}$ & - & - \\
w/. Vista$^*$~\cite{gao2024vista} & $C$ & 76.62$\mathrm{\tiny{\hi{+2.32}}}$ & 37.71$\mathrm{\tiny{\hi{+1.71}}}$ &  34.04$\mathrm{\tiny{\hi{+1.16}}}$ & 38.60$\mathrm{\tiny{\hi{+0.79}}}$  \\  
w/. MagicDrive~\cite{gao2023magicdrive} & $C$ & 79.56$\mathrm{\tiny{\hi{+5.26}}}$ & 40.34$\mathrm{\tiny{\hi{+4.34}}}$ & 35.40$\mathrm{\tiny{\hi{+2.52}}}$ & 39.76 $\mathrm{\tiny{\hi{+1.95}}}$ \\
\rowcolor{gray!10}w/. Ours & $C$ &  \textbf{81.69}$\mathrm{\tiny{\hi{+7.39}}}$ & \textbf{41.62}$\mathrm{\tiny{\hi{+5.62}}}$  & \textbf{36.50}$\mathrm{\tiny{\hi{+3.62}}}$ & \textbf{41.22}$\mathrm{\tiny{\hi{+3.41}}}$  \\ \midrule

 Baseline-M~\cite{liu2023bevfusion} & $C\&L$  & -  &   - & 65.40 & 69.59 \\
\rowcolor{gray!10}w/. Ours-M &  $C\&L$  & -  & -  &  \textbf{68.53}$\mathrm{\tiny{\hi{+3.13}}}$ & \textbf{72.20}$\mathrm{\tiny{\hi{+2.61}}}$ \\
\bottomrule
\end{tabular}
}
\vspace{-8pt}
\caption{
\textbf{Comparison about training support for BEV segmentation} (Baseline as CVT~\cite{zhou2022cross}) and \textbf{3D object detection} (Baseline as BEVFusion~\cite{liu2023bevfusion}).
}
\vspace{-23pt}
\label{tab_down2}
\end{center}
\end{table}

\begin{figure}[!t]
    \centering
    \includegraphics[width=1.0\linewidth]{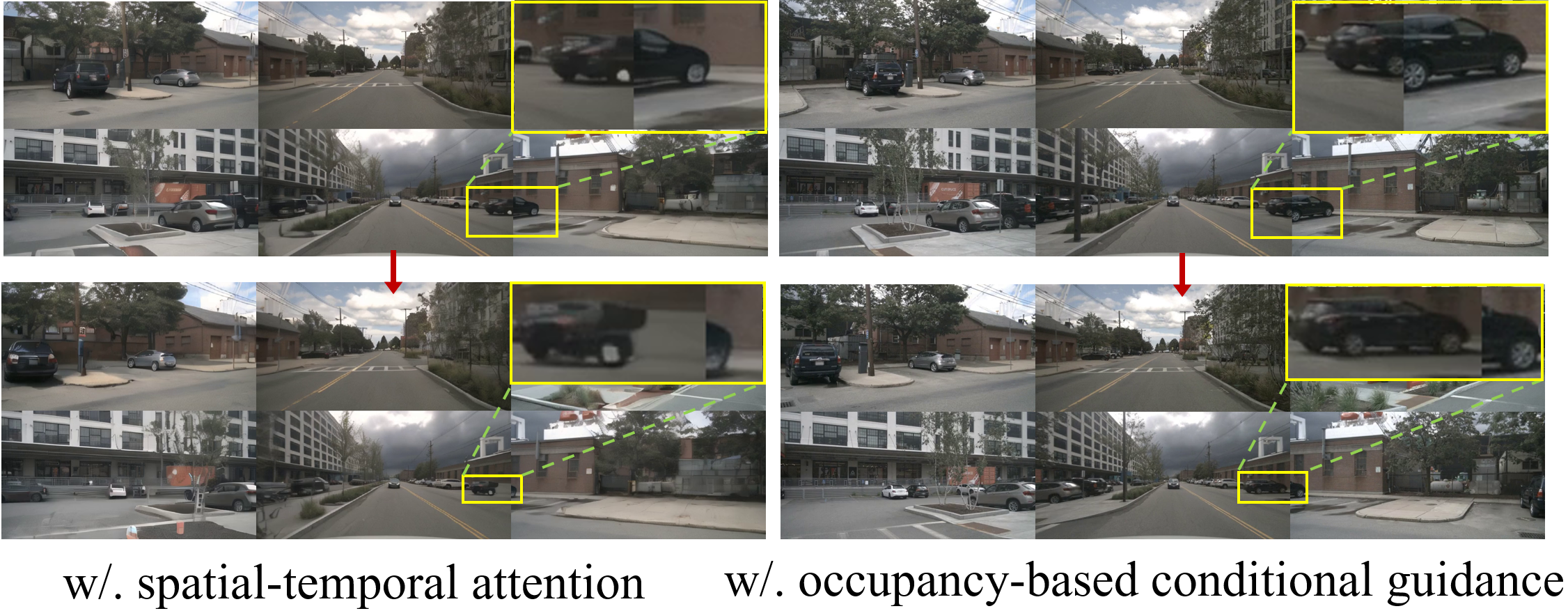}
    \vspace{-21pt}
    \caption{\textbf{Visualization} of the effect of occupancy-based conditional guidance. Our approach significantly enhances cross-view consistency, resulting in high-fidelity structures, while spatial-temporal attention fails to achieve similar results.}
    \label{fig_video_ablation}
    \vspace{-12pt}
\end{figure}

\noindent\textbf{Effect of Designs in Video Generation Model.}
As shown in Tab.~\ref{table_ar_video} and Fig.~\ref{fig_video_ablation}, we conduct extensive ablation studies to validate the effectiveness of the video generation model components.
For the setting of `w/. Spatial-temporal Attention', we remove the occupancy-based rendering maps and employ spatial-temporal attention as~\cite{wu2023tune,gao2023magicdrive}.
Our results indicate that the occupancy-based semantic and geometric rendering maps are more effective in enhancing video generation quality.
Moreover, as illustrated in Fig.~\ref{fig_video_ablation}, our strategy markedly improves the cross-view consistency with high-fidelity structures.
The rendered semantic and depth maps from the semantic occupancy benefit the fidelity of image and video generation significantly with detailed prior guidance, boosting 34.66\% and 31.15\% in terms of FVD, respectively. 
The geometric-aware noise prior improves the video generation performance obviously, reducing FVD from 87.52 to 70.52.

\noindent\textbf{Effect of Designs in LiDAR Generation Model.}
The ablation studies on the proposed LiDAR generation model are illustrated in Tab.~\ref{table_ar_lidar}. For the setting of `w/o. Sparse UNet', we replace the Sparse UNet with a single layer of submanifold convolution~\cite{graham2017submanifold}.
As we can see, the Sparse UNet efficiently improves the generation performance (-25.77\% in JSD) with a relatively small memory increment (+1.63\%). 
To implement the setting of `w/o. Sparse Sampling', we replace it with uniform sampling following NeuS~\cite{wang2021neus}. Our sparse sampling strategy brings a significant computational consumption decrease (-58.94\%) with better performance (-4.00\% in JSD). 
Moreover, the ray-dropping head reduces MMD and JSD by 25.92\% and 25.00\%, respectively. We attribute such obvious improvement to the capability of realistic LiDAR ray-dropping phenomenon simulation.

\begin{table}[!t]
\vspace{-0pt}
\centering
\renewcommand\tabcolsep{13.5    pt}
\resizebox{1.00\columnwidth}{!}{
\begin{tabular}{l|ccc}
\toprule
        Method & mIoU $\uparrow$  & F3D $\downarrow$ & MMD $\downarrow$ \\
         \midrule 
    \rowcolor{gray!10}Ours & \textbf{19.44} & \textbf{158.55} & \textbf{10.60}   \\ \midrule
    w/o. VAE 3D Axial Attention & 18.77 & 167.91 & 17.10 \\
    w/o. DiT Temporal Attention & 17.63 & 176.74 & 11.35 \\
    w/o. DiT Spatial Attention & 10.29 & 261.03& 18.59\\
    \bottomrule
    \end{tabular}%
}
\vspace{-8pt}
\caption{\textbf{Ablation} for designs in the occupancy generation model.
} 
\label{table_ab_occ_1}
\vspace{-6pt}
\end{table}

\begin{table}[!t]
\vspace{-0pt}
\centering
\renewcommand\tabcolsep{23.2pt}
\resizebox{1.00\columnwidth}{!}{
\begin{tabular}{l|ccc}
\toprule
Method & FID$\downarrow$ & FVD$\downarrow$ \\ \midrule
\rowcolor{gray!10}Ours  & {\textbf{6.12}} & {\textbf{70.52}}   \\ \midrule
 w/. Spatial-temporal Attention & 12.72 & 110.87 \\ 
 w/o. Rendered Semantic Map & 11.72 & 107.92 \\
 w/o. Rendered Depth Map   &10.17  & 102.42 \\ 
 w/o. Depth-based Noise Prior &7.23 & 87.52 \\
 \bottomrule
\end{tabular}
}
\vspace{-8pt}
\caption{\textbf{Ablation} for designs in the video generation model.
} 
\label{table_ar_video}
\vspace{-6pt}
\end{table}

\begin{table}[!t]
\vspace{-0pt}
\centering
\renewcommand\tabcolsep{2.7pt}
\resizebox{1.00\columnwidth}{!}{
\begin{tabular}{l|cccc}
\toprule
Method & MMD ($10^{-4}$) $\downarrow$ &JSD $\downarrow$ & Time (s)$\downarrow$ & Memory (GB)$\downarrow$   \\ \midrule
\rowcolor{gray!10}Ours  & \textbf{1.53} & \textbf{0.072} & {0.25} & 6.84 \\ \midrule
w/o. Sparse UNet  & 2.88 & 0.097 & \textbf{0.21} & 6.73 \\ 
w/o. Sparse Sampling & 1.69 & 0.075 & 0.30 & 16.66\\ 
w/o. Ray-dropping Head & 3.25 & 0.100 & 0.25 & \textbf{5.05} \\
\bottomrule
\end{tabular}
}
\vspace{-8pt}
\caption{
\textbf{Ablation} for designs in the LiDAR generation model.
} 
\label{table_ar_lidar}
\vspace{-6pt}
\end{table}

\vspace{-2pt}
\section{Conclusion and Future Work} 
In this paper, we introduce UniScene, a unified framework designed to generate high-fidelity, controllable, and annotated data for autonomous driving applications. By decomposing the complex scene generation task into two hierarchical steps, UniScene progressively produces semantic occupancy, video, and LiDAR data. Extensive experiments show that UniScene surpasses current SOTAs in all three data types and enhances extensive downstream tasks.

\noindent\textbf{Limitations and Future Work.} Unifying a comprehensive system that integrates multiple generation tasks is challenging and resource-intensive. Exploring methods to optimize the system for lightweight deployment is a promising avenue for future research. Additionally, extending the system to relevant domains such as embodied intelligence and robotics presents valuable opportunities for further advancement.

\newpage
\appendix
\twocolumn[{
\centering
\vspace{20pt}
\section*{\Large \centering Supplementary Material for UniScene}
 \vspace{30pt}
 }]


\section{Problem Definition and Distinctiveness}

\subsection{Controllable Occupancy Generation} 
Previous research on uncontrollable occupancy generation~\cite{lee2023diffusion, lee2024semcity, liu2023pyramid} has primarily focused on the creation of static scenes, which limits their applicability to dynamic scenarios due to a lack of controllability. In contrast, our approach introduces controllable generation $\mathbb{P}(\texttt{Occ} | \texttt{BEV})$, effectively incorporating temporal information into the process. Here $\texttt{BEV}$ refers to input Bird’s Eye View (BEV) scene layouts and $\texttt{Occ}$ denotes the corresponding generated semantic occupancy, respectively.\looseness=-1

\subsection{Conditional Video and LiDAR Generation} 
Existing methods for video~\cite{wang2023drivedreamer,zhao2024drivedreamer,gao2023magicdrive,wang2023driving,wen2024panacea} and LiDAR~\cite{zyrianov2022learning,ran2024towards,zyrianov2024lidardm} generation typically produce data directly from coarse scene layouts (e.g., BEV maps, 3D bounding boxes). However, these approaches often fail to accurately capture the intricate distributions inherent in driving scenes, leading to suboptimal performance.
In contrast, our proposed UniScene framework addresses these limitations by decomposing the complex generation task into a hierarchy centered around occupancy. This is formally expressed as:
\begin{equation}
\begin{aligned}
\mathbb{P}(\texttt{Vid}, \texttt{Lid} | \texttt{BEV}) &= \mathbb{P}(\texttt{Vid}, \texttt{Lid} | \texttt{Occ}) \cdot \mathbb{P}(\texttt{Occ} | \texttt{BEV}),
\end{aligned}
\end{equation}
where $\texttt{Vid}$, and $\texttt {Lid}$ denote the generated video, and LiDAR data, respectively.
By leveraging occupancy priors, our method alleviates the learning burden and more accurately captures the underlying distributions for performance enhancement, as demonstrated in Fig. 1(a) of the main paper.

\section{More Related Works and Discussions}

\subsection{Semantic Occupancy Generation} SemCity ~\cite{lee2024semcity} proposes a 3D semantic scene generation approach with a triplane diffusion framework. PyramidOcc~\cite{liu2023pyramid} generates large-scale 3D semantic scenes using a coarse-to-fine paradigm with pyramid discrete diffusion~\cite{austin2021structured}. These methods mainly focus on unconditional and static 3D scene generation.  More recent work~\cite{zhang2024urban} is capable of controlling the generation of 3D scenes through BEV maps. However, it remains confined to static scenes. OccSora ~\cite{wang2024occsora} generates temporal 3D scene sequences with a diffusion transformer (DiT)~\cite{peebles2023scalable}. Due to the high compression ratio of occupancy in its designed VQVAE, the reconstruction performance is suboptimal, which to some extent leads to unsatisfactory generation quality. Moreover, it lacks the capability of precisely controlling the generated results. To address the issues in these works, we propose our occupancy generation model, enabling controllable generation of temporal 3D scene sequences with high fidelity while effectively maintaining temporal consistency.\looseness=-1

\subsection{Driving Video Generation} 
Vista~\cite{gao2024vista} builds on the architecture of Stable Video Diffusion (SVD)~\cite{blattmann2023stable} to enable single-view driving video generation with various action controls. However, this method is constrained by its inability to produce multi-view videos and its lack of alignment between the generated outputs and ground-truth labels, which hinders its utility for training downstream tasks.
WoVoGen~\cite{lu2023wovogen} presents a world model that predicts future videos and occupancy based on past observations. In this model, occupancy grids are compressed into low-dimensional features~\cite{radford2021learning}, potentially leading to suboptimal generation results (see the seventh row of Table 4 in the main paper).
Recent work of SyntheOcc~\cite{li2024syntheocc} synthesizes multi-view images in driving scenarios using occupancy-based multi-plane semantic images. Nevertheless, this method neglects the explicit geometric information within the semantic occupancy grids and necessitates substantial manual intervention for occupancy grid editing.
Our proposed UniScene aims to jointly render semantic and depth maps from the semantic occupancy, thereby providing detailed prior information. Moreover, UniScene simplifies geometric editing by using BEV maps as scene layouts.\looseness=-1

\subsection{LiDAR Point Cloud Generation} 
Pioneering works in LiDAR generation~\cite{zyrianov2022learning,hahner2022lidar,xiong2023learning,xiong2023ultralidar,zyrianov2024lidardm,ran2024towards} utilized GAN or diffusion models to produce LiDAR point clouds. LiDARGen~\cite{zyrianov2022learning} adopts an equirectangular view image as a structured representation of LiDAR point clouds and uses a score-based diffusion model for point cloud generation. Nevertheless, the 2.5D representation may constrain its capacity to accurately generate 3D geometries of real-world objects. Furthermore, LiDARGen~\cite{zyrianov2022learning} applies the diffusion process directly to LiDAR points rather than in the latent space, significantly slowing down the inference process.
UltraLiDAR~\cite{xiong2023ultralidar} voxelized LiDAR points and transformed them into a BEV representation, employing VQVAE to learn a compact 3D representation of LiDAR points and a generative transformer for LiDAR points generation. However, using BEV as the representation overlooks the fine geometric details in the LiDAR data, potentially impacting the generation quality.
LiDM~\cite{hahner2022lidar} employs range maps as the representation for LiDAR data, integrating curve-wise compression, patch-wise encoding, and point-wise coordinate supervision in VQVAE to enhance its geometric representation capabilities. 
However, the use of range map representation can result in a loss of structured LiDAR point information. In this work, we propose to facilitate high-fidelity LiDAR point generation by leveraging fine-grained priors from semantic occupancy grids.\looseness=-1

\section{More Implementation Details}

\subsection{BEV Editing Scheme}
The modification of BEV layouts for the controllable generation of semantic occupancy, videos, and LiDAR point clouds represents a significant application in the creation of out-of-distribution (OOD) data. To accomplish this, our editing scheme is structured as follows:\looseness=-1

\begin{itemize}
\item Modify the initial BEV layout, $\mathbf{B}_\text{ori}$, to produce a revised layout, $\mathbf{B}_\text{new}$, incorporating specific alterations (e.g., the removal of a vehicle).
\item Employ DDIM Inversion \cite{mokady2023null} to transform the original occupancy, $\mathbf{O}_\text{ori}$, into a noise latent, $\epsilon_\text{ori}$, while being guided by the original BEV layout, $\mathbf{B}_\text{ori}$.
\item Construct the updated occupancy, $\mathbf{O}_\text{new}$, via denoising diffusion, utilizing the modified BEV layout, $\mathbf{B}_\text{new}$, as a conditioning factor and $\epsilon_\text{ori}$ as the starting noise.
\item With the newly generated occupancy, $\mathbf{O}_\text{new}$, serve as the addition condition, produce the corresponding video, $\mathbf{V}_\text{new}$, and LiDAR data, $\mathbf{L}_\text{new}$, with detailed prior guidance.
\end{itemize}

\begin{figure}[!t]
    \centering
    \includegraphics[width=1\linewidth]{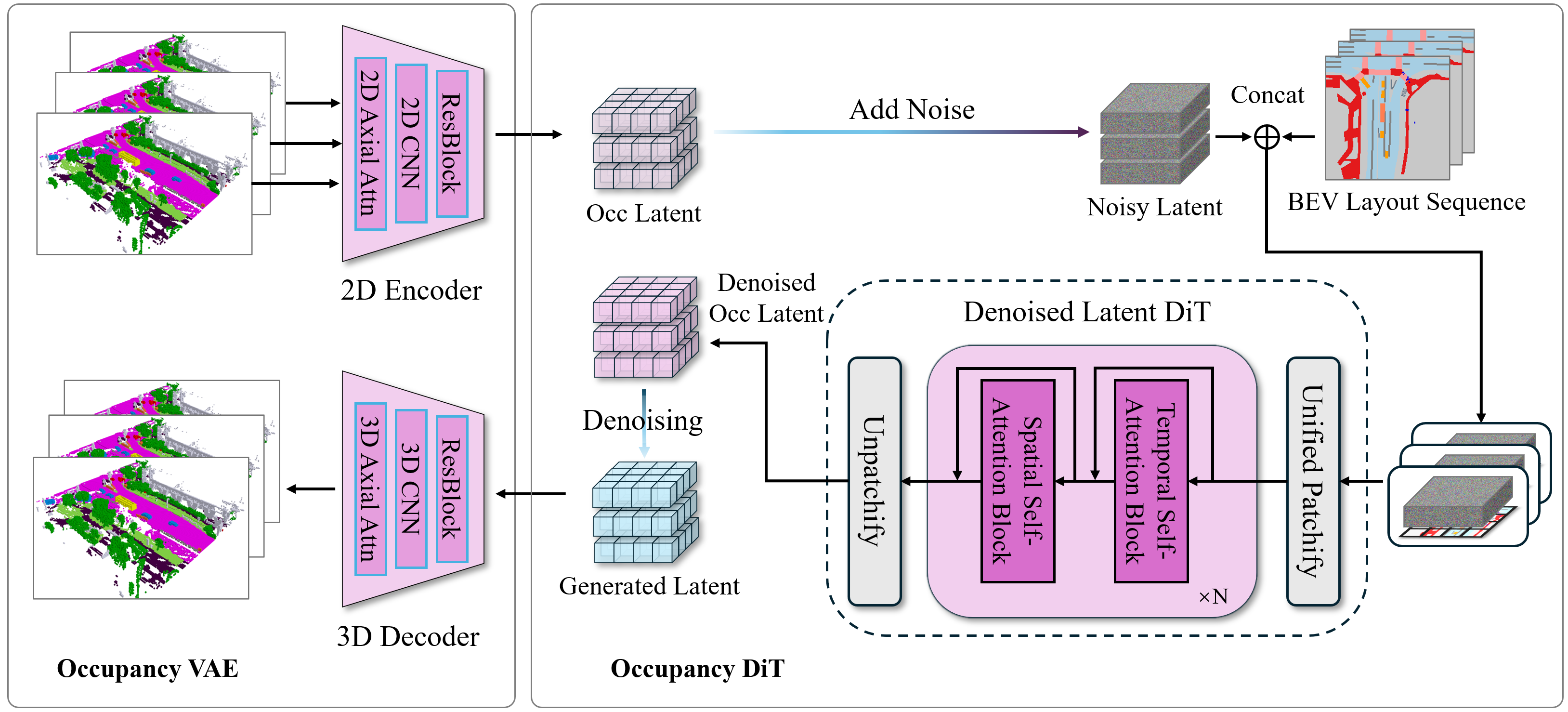}
    \caption{The architecture of the occupancy generation model, which consists of two main components: the Occupancy VAE and the Occupancy DiT. The Occupancy VAE includes a 2D encoder, leveraging ResBlock, 2D CNNs, and axial attention, and a 3D decoder to reconstruct the input data. The Occupancy DiT applies a denoising diffusion process using temporal and spatial self-attention blocks to generate denoised latent representations. Noisy latents, combined with BEV layout sequences, are processed to create unified patchified outputs, enabling robust occupancy generation.}
    \label{occgen_framework}
\end{figure}

\subsection{Occupancy Generation Model}
As illustrated in \cref{occgen_framework}, the occupancy generation model is composed of a VAE and a DiT, which produces semantic occupancy with a compressed latent space.

\begin{table}[!t]
\vspace{-0pt}
\begin{center}
\scriptsize
\vspace{-0pt}
\renewcommand\tabcolsep{7.6pt}
	\centering
   \resizebox{0.99\linewidth}{!}{
\begin{tabular}{l|c|c|c}
\toprule Method  & 
\begin{tabular}[c]{@{}c@{}}Compression\\Ratio\end{tabular} $\uparrow$ & mIoU $\uparrow$ & IoU $\uparrow$ \\ \midrule
OccLLama (VQVAE)~\cite{wei2024occllama} & 8 & \underline{75.2}& 63.8\\
OccWorld (VQVAE)~\cite{zheng2023occworld}  & 16 & 65.7 & 62.2 \\
OccSora (VQVAE)~\cite{wang2024occsora} & \textbf{512} & 27.4& 37.0\\
\midrule

Ours(VQVAE) & \underline{32} & 59.8 & 58.2\\
\rowcolor{gray!10} Ours (VAE)  &  \underline{32}  & \textbf{92.1} &  \textbf{87.0} \\
 Ours(VQVAE) & \textbf{512} & 55.8 & 56.8\\
\rowcolor{gray!10} Ours (VAE) & \textbf{512} & 72.9 & \underline{64.1}\\
\bottomrule
\end{tabular}
 }
 \vspace{-8pt}
 \caption{Comparison of VQVAE and VAE performance on occupancy reconstruction, with compression ratios calculated based on the methodology from OccWorld~\cite{zheng2023occworld}. The results demonstrate the clear superiority of our occupancy VAE over VQVAE under the same architectural design, achieving significantly higher mIoU and IoU scores across various compression ratios.\looseness=-1 
 }
\label{tab_occ_rec_appendix}
\end{center}
\end{table}

\noindent\textbf{Occupancy VAE.}  
We leverage the Occupancy VAE encoder to transform a 3D semantic occupancy $\mathbf{O} \in \mathbb{R}^{H \times W \times D}$ within an occupancy sequence into a BEV representation $\mathbf{\hat{O}} \in \mathbb{R}^{H \times W \times DC'}$ by assigning each category a learnable class embedding $C'$. A 2D CNN encoder with a 2D axial attention layer is utilized to extract a continuous latent feature with down-sampled resolution $\mathbf{Z}_\text{occ} \in \mathbb{R}^{C \times h \times w}$, where $h = \frac{H}{d}$ and $w = \frac{W}{d}$, with $d$ being the down-sampling factor.  

The VAE decoder reconstructs the latent feature sequence $\mathbf{z}_\text{occ}^\text{seq} \in \mathbb{R}^{T \times C \times h \times w}$. A 3D CNN network with a 3D axial attention layer is employed to up-sample the latent feature sequence to a BEV representation occupancy sequence $\mathbf{\hat{O}}^\text{seq} \in \mathbb{R}^{T \times H \times W \times DC'}$. This sequence is then reshaped to $\mathbb{R}^{THW \times D \times C'}$ and processed through a dot product with the class embeddings to obtain the logits scores. During training, the logits scores and one-hot labels are used to compute the cross-entropy loss and Lovász-softmax loss~\cite{berman2018lovasz}. In the inference phase, the final reconstructed occupancy sequence $\mathbf{O}^\text{seq} \in \mathbb{R}^{T \times H \times W \times D}$ is determined by taking the $\texttt{argmax}$ of the logits. We provide more occupancy reconstruction results in Tab.~\ref{tab_occ_rec_appendix} to illustrate the superiority of our occupancy VAE design with continuous compression.
The results demonstrate the clear superiority of our occupancy VAE over VQVAE under the same architectural design, which achieves significantly higher mIoU and IoU scores across various compression ratios.\looseness=-1

\begin{figure}[!t]
    \centering
\includegraphics[width=1.00\linewidth]{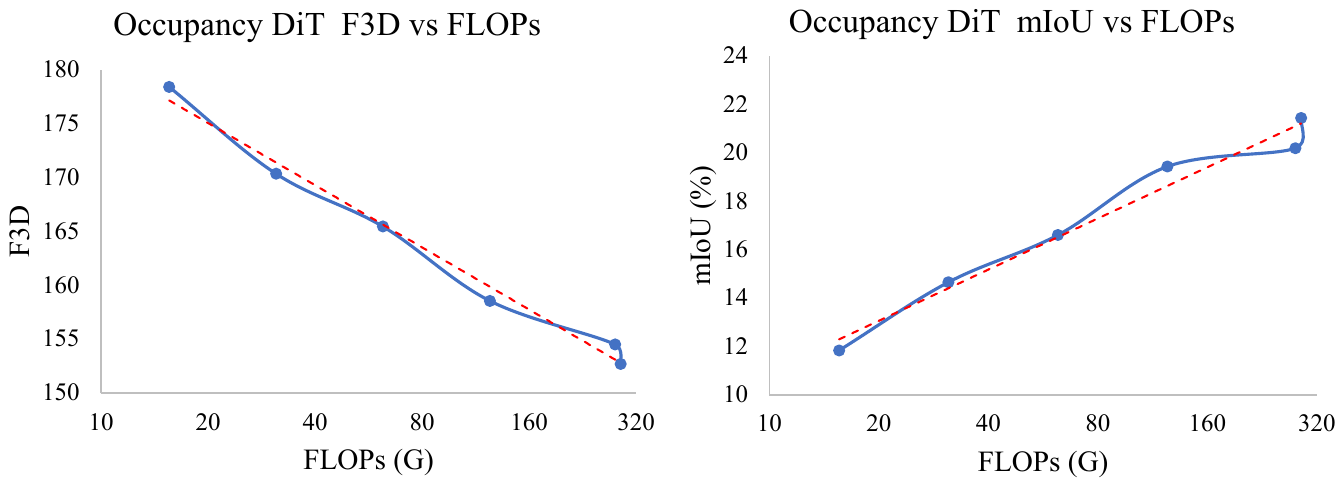}
    \caption{Scalability analysis of Occupancy DiT, illustrating how performance scales with increasing computational cost (FLOPs). The left plot shows a consistent improvement in F3D performance as FLOPs increase, while the right plot highlights the scalability of mIoU, which also improves steadily with higher computational resources, demonstrating the model's capacity to leverage increased computation for better results.}
    \label{fig_DiT_scaling}
\end{figure}

\noindent\textbf{Occupancy DiT.}
As shown in \cref{occgen_framework}, we employ an Occupancy DiT to denoise occupancy latent sequence features from noisy occupancy latents and BEV layout sequences. To align the BEV layout sequence with the occupancy latent features, we introduce a unified patchify module. Specifically, the BEV layout at time step $i$ is down-sampled into $\mathbf{B}_{down}^{i} \in \mathbb{R}^{(C_b) \times h \times w} $ to match the shape of the latent feature and then concatenated with the latent $\mathbf{Z}_\text{occ}^{i} \in \mathbb{R}^{(C_o) \times h \times w}$. A unified patch embedder transforms the concatenated latent $\mathbf{Z}_\text{cat} \in \mathbb{R}^{(C_o+C_b) \times h \times w}$ into unified latent tokens $\mathbf{Z} \in \mathbb{R}^{L \times E_d}$, where $L$ is the number of patches and $E_d$ is the embedding dimension.\looseness=-1

The backbone of Occupancy DiT is the Spatial-Temporal Latent Diffusion Transformer, comprising stacked spatial and temporal blocks. Spatial blocks aggregate features across different positions within the same latent, whereas temporal blocks capture features across different latent frames at the same position. Additionally, 2D positional embeddings and 1D temporal embeddings are used to account for relative relationships. The output of the backbone, with dimensions $\mathbb{R}^{T \times L \times E_d}$, is passed through an unpatchify layer to yield a denoised occupancy latent sequence of size $\mathbb{R}^{T \times H \times W \times D}$.
During the training phase, we randomly drop the BEV layout condition with a probability of 0.1, enabling the diffusion model to learn unconditional generation, which is essential for occupancy editing. In the sampling phase, the classifier-free guidance scale is set to 1.0 by default.

\begin{table}[!t]
\vspace{-0pt}
\begin{center}
\scriptsize
\vspace{-0pt}
\renewcommand\tabcolsep{3.6pt}
	\centering
   \resizebox{0.99\linewidth}{!}{
    \begin{tabular}{c|cccc|cccc}
    \toprule
    \multicolumn{1}{c|}{\multirow{2}[4]{*}{Method}} & \multicolumn{4}{c|}{mIoU $\uparrow$} & \multicolumn{4}{c}{IoU $\uparrow$} \\
\cmidrule{2-9}    \multicolumn{1}{c|}{} & 1s    & 2s    & 3s    & Avg   & 1s    & 2s    & 3s    & Avg \\
    \midrule
    OccWorld \cite{zheng2023occworld} & 25.75 & 15.14 & 10.51 & 17.13 & 34.63 & 25.07 & 20.19 & 26.63 \\
    OccLLama \cite{wei2024occllama}& 25.05 & 19.49 & 15.26 & 19.93 & 34.56 & 25.83 & 24.41 & 29.17 \\
    \midrule
    \rowcolor{gray!10} Ours-Fore. (1 ref)& 30.93 & 24.87 & 20.75 & 27.33 & 35.15 & 31.79 & 29.24 & 31.62 \\
    \rowcolor{gray!10} Ours-Fore. (2 ref) & 35.37 & 29.59 & 25.08 & 31.76 & 38.34 & 32.70  & 29.09 & 34.84 \\
    \bottomrule
    \end{tabular}%
    }
 \caption{Comparison of occupancy forecasting performance, where Ours-Fore. (1 ref) and Ours-Fore. (2 ref) represent our forecasting model conditioned on 1 and 2 reference occupancy frames, respectively. The results highlight significant improvements in both the mIoU and IoU metrics over baseline methods (OccWorld and OccLLama) across different time horizons (1s, 2s, 3s), demonstrating the effectiveness of our approach in generating accurate and consistent occupancy predictions.\looseness=-1 }
\label{occ_fore_tab}
\end{center}
\end{table}

\noindent\textbf{Occupancy Forecasting Variant.}
To facilitate comparison with other occupancy forecasting works \cite{zheng2023occworld, wei2024occllama}, the occupancy generation model is adapted into a generative forecasting model, which predicts $T_f$ future frames based on $T_c$ conditional frames. 
Specifically, during the training phase, the conditional occupancy latent frames are concatenated directly with the BEV layouts without the addition of noise. The unified latent tokens for both $T_c$ and $T_f$ frames are then fed into the DiT backbone, following the same procedure as in the generation model. The model outputs denoised occupancy latent frames for both $T_c$ and $T_f$ frames, while the loss is computed only on the $T_f$ frames.
In the inference phase, the $T_f$ frames are initialized with pure noise, while the $T_c$ frames are initialized with conditional occupancy latents sampled from the Occupancy VAE. The future occupancy frame number $T_f$ is set to $6$ to align with previous works~\cite{zheng2023occworld, wei2024occllama}. To improve computational efficiency and reduce dependence on the conditional frames, we set $T_c = 1$ or $T_c = 2$ instead of $T_c = 5$ in previous works~\cite{zheng2023occworld, wei2024occllama}. The results of the forecasting model are presented in~\cref{occ_fore_tab}.
As we can see from the table, our method effectively surpasses other works across
different time horizons (1s, 2s, 3s), demonstrating the effectiveness of our approach in generating accurate and consistent occupancy generation. 
Moreover, we visualize the scalability of the Occupancy DiT in Fig.~\ref{fig_DiT_scaling} to demonstrate the potential of our proposed method.
The performance improves steadily with higher computational resources, which demonstrates the model’s capacity to leverage increased
computation for better results as discussed in~\cite{Peebles2022DiT}.\looseness=-1

\subsection{Video Generation Model}
\begin{figure}[!t]
    \centering
    \includegraphics[width=1.0\linewidth]{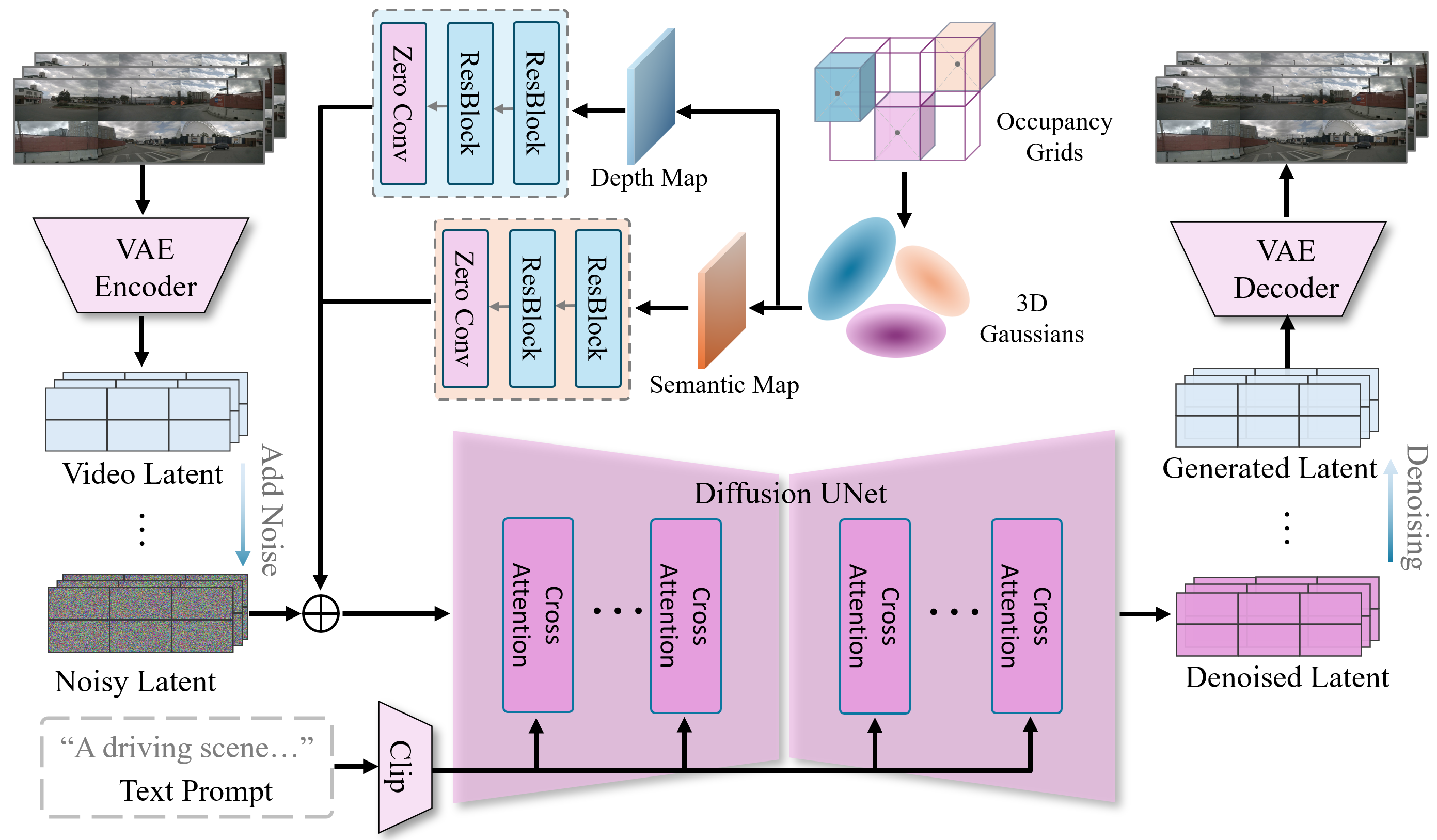}
    \caption{The architecture of the video generation model, which combines video latent representations with textual prompts to guide video generation. The VAE encoder extracts video latent features, which are combined with noise before passing through the Diffusion UNet. The Diffusion UNet employs cross-attention mechanisms to integrate textual guidance and refine the latent representation. Occupancy grids serve as the basis for generating depth maps and semantic maps, which act as guidance for the video generation process. The VAE decoder reconstructs the video from the denoised latent, producing realistic outputs guided by the input prompts.}
    \label{videonet}
\end{figure}

As shown in Fig.~\ref{videonet}, the video generation model combines video latent representations with occupancy grids and textual prompts to guide video generation. The VAE encoder extracts video latent features, which are combined with noise before passing through the Diffusion UNet.

\noindent\textbf{Preliminaries with Stable Video Diffusion.}
The Stable Video Diffusion (SVD)~\cite{blattmann2023stable} is a latent diffusion model specifically designed for image-to-video (I2V) generation. To increase sampling flexibility, SVD utilizes a continuous-timestep formulation. The model converts data samples $x$ into noise $n$ through a diffusion process, where $p(n|x)$ follows a Gaussian distribution $\mathcal{N}(x, \sigma^2 I)$. New samples are then generated by gradual denoising the latent space from Gaussian noise until $\sigma_0 = 0$.
SVD generates videos by processing a series of noisy latent representations, with the generation being guided by a conditional image. The latent representation of this image is channel-wise concatenated to the input, acting as a reference for the content creation.

\noindent\textbf{Preliminaries with Gaussian Splatting.}
The Gaussian Splatting~\cite{gaussian_splatting} approach defines a set of 3D Gaussian primitives as $\mathcal{G} = \{G_i\}_{i=1}^{N}$, where each primitive $G_i$ contains attributes of position $\boldsymbol{\mu}_i$, color ${c}_i$, opacity ${\alpha}_i$, rotation matrix $\mathbf{R}_i$, and scale matrix $\mathbf{S}_i$. The associated 3D covariance matrix $\mathbf{\Sigma}_i$ for each primitive is derived as:
\begin{equation}
\mathbf{\Sigma}_i= \mathbf{R}_i \mathbf{S}_i \mathbf{S}_i^{\mathbf{T}} \mathbf{R}_i^{\mathbf{T}},
\end{equation}
when transformed by a viewing transformation 
$\mathbf{W}$, the covariance matrix in camera coordinates $\mathbf{\Sigma}^{'}_{i}$ is computed as:

\begin{equation}
\mathbf{\Sigma}^{'}_{i}= \mathbf{J} \mathbf{W} \mathbf{\Sigma}_i \mathbf{W}^{\mathbf{T}} \mathbf{J}^{\mathbf{T}},
\end{equation}
where $\mathbf{J}$ is the Jacobian of the affine approximation of the projective transformation. 
This process yields a $2 \times 2$ covariance matrix, consistent with the previous work~\cite{gaussian_splatting}. 
For a given projected 3D Gaussian center $\boldsymbol{\mu} \in \mathbb{R}^{2 \times 1}$ and a point $\mathbf{x} \in \mathbb{R}^{2 \times 1}$ in camera coordinates, the opacity ${\alpha}^{\prime}$ of projected 2D Gaussian is formulated as:
\begin{equation}
{\alpha}^{\prime}={\alpha} \exp \left(-\frac{1}{2}(\mathbf{x}-\boldsymbol{\mu})^T\left(\mathbf{\Sigma}^{\prime}\right)^{-1}(\boldsymbol{x}-\boldsymbol{\mu})\right).
\end{equation}

The accumulated color map $\mathbf{C}$ is obtained via tile-based rasterization:
\begin{equation}
\mathbf{C}=\sum_{i \in \mathcal{N}} {c}_i \alpha_i^{\prime} \prod_{j=1}^{i-1}\left(1-\alpha_j^{\prime}\right).
\end{equation}

\noindent\textbf{Gaussian-based Joint Rendering.}
In this work, to generate depth and semantic maps, we start by transforming an input semantic occupancy grid, with dimensions $\mathbb{R}^{H \times W \times D}$, into a set of 3D Gaussian primitives $\mathcal{G}$. Each primitive $G_i$ is assigned an additional attribute: the semantic label $s_i$. The position of each 3D Gaussian is initialized using the center of the corresponding occupancy grid cell, its scale is set according to the size of the grid cell, and its semantic attribute is initialized with the semantic label of the grid cell.
To improve the effectiveness of the 2D conditions, we render depth and semantic maps that align with the camera's viewpoint. Specifically, given the semantic label $s_i$ and depth value $d_i$, the depth map $\mathbf{D}$ and the semantic map $\mathbf{S}$ are rendered in a manner analogous to the color rendering:
\begin{equation} 
\mathbf{D}=\sum_{i \in {N}} {d}_i \alpha_i^{\prime} \prod_{j=1}^{i-1}\left(1-\alpha_j^{\prime}\right),
\end{equation}
\vspace{-10pt}
\begin{equation} 
\mathbf{S}=\texttt{argmax} ( \sum_{i \in {N}} \texttt{onehot}({s}_i) \alpha_i^{\prime} \prod_{j=1}^{i-1}\left(1-\alpha_j^{\prime}\right)).
\end{equation}

As discussed in Section 3.2 of the main paper, we incorporate an encoder branch inspired by ControlNet~\cite{zhang2023adding} to condition the rendering maps. The specifics of this conditioning are depicted in Fig.~\ref{videonet}. The encoder branch consists of two modules, each designed for either depth map conditioning or semantic map conditioning, with similar architectures. Each conditioning module comprises two ResNet blocks followed by a zero convolution, as recommended by ControlNet~\cite{zhang2023adding}, to preserve the original functionality of the diffusion model.

\begin{figure}[!t]
    \centering
    \includegraphics[width=1\linewidth]{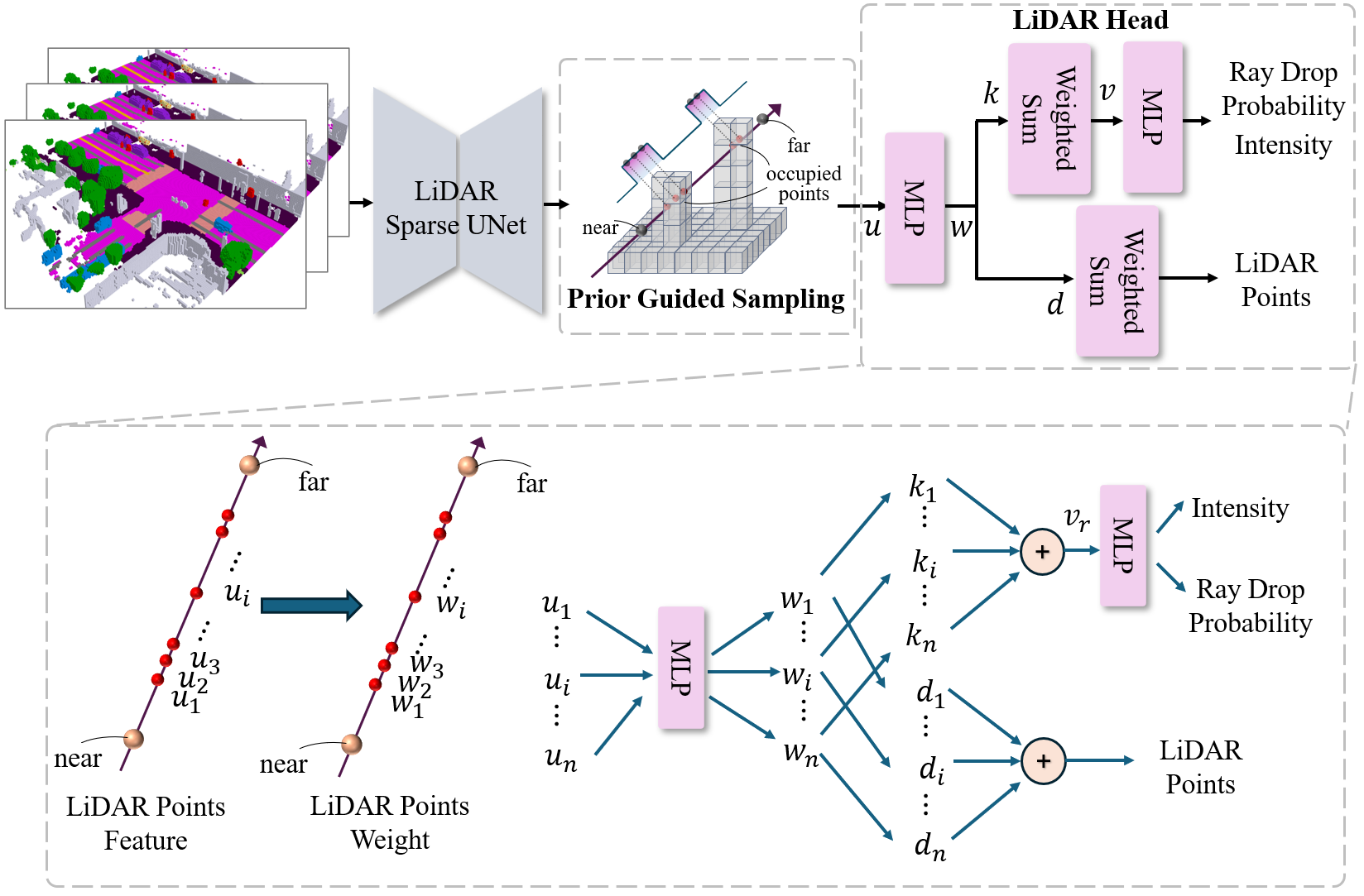}
    \caption{The architecture of the LiDAR generation model, which integrates a LiDAR Sparse UNet and a Prior Guided Sampling mechanism. The LiDAR Sparse UNet processes input occupancy data to extract spatial features, while the Prior Guided Sampling generates LiDAR points by sampling along rays based on the spatial structure. The LiDAR Head employs weighted summation and MLP modules to calculate the intensity, ray drop probabilities, and final positions of the LiDAR points. The process involves assigning weights to the generated LiDAR point features, refining them through an MLP, and aggregating the results to produce high-quality LiDAR outputs with realistic point distributions.}
    \label{supple_lidar}
\end{figure}

\subsection{LiDAR Generation Model}
As illustrated in Fig.~\ref{supple_lidar}, 
the LiDAR generation model comprises a LiDAR Sparse UNet and a Prior Guided Sampling mechanism. The LiDAR Sparse UNet processes input occupancy data to extract spatial features, whereas the Prior Guided Sampling mechanism generates LiDAR points by sampling along rays according to the spatial structure. The LiDAR Head utilizes weighted summation and multilayer perceptron (MLP) modules to compute the intensity, ray drop probabilities, and final positions of the LiDAR points. This process involves assigning weights to the generated LiDAR point features, refining these features through an MLP, and aggregating the results to produce high-quality LiDAR outputs with realistic point distributions.

Specifically, for $n$ points on a LiDAR ray, the feature of the $i$-th point, ${u}_i$, is sampled from the output of the LiDAR Sparse UNet with the occupancy-based prior guidance. This feature is then fed into an MLP to predict the corresponding Signed Distance Function (SDF) value. The weight for the point, $w_i$, is subsequently computed and used to determine the ray depth $d_i$ of the generated LiDAR points according to Eq. 8 and Eq. 9 in the main paper. Additionally, the ray feature, ${v}_r$, is obtained by performing a weighted sum of the features of all points along the ray, using their respective weights, that is, 
\begin{equation}
    v_r =  \sum_{i=1}^{n} k_i  =  \sum_{i=1}^{n} w_i \cdot u_i.
\end{equation}
Finally, ${v}_r$ is passed through another MLP to simultaneously predict the intensity and drop probability of the LiDAR ray.\looseness=-1

\section{Datasets}
The NuScenes benchmark~\cite{caesar2020nuscenes} is widely used in autonomous driving research. To enrich this dataset with more detailed annotations, the NuScenes-Occupancy dataset~\cite{wang2023openoccupancy} has introduced dense semantic occupancy labels. This dataset encompasses comprehensive LiDAR sweep data across 850 scenes, comprising 34,000 key-frames at a frame rate of 2 Hz, each annotated with one of 17 semantic categories. Following the methodology described in~\cite{wang2023openoccupancy}, our study allocates 28,130 frames for the training set and 6,019 frames for validation.\looseness=-1

However, the 2 Hz data alone does not meet our requirements. To enhance our results and ensure a fair comparison with previous works~\cite{gao2023magicdrive, wang2023driving,wen2024panacea,swerdlow2024streetview,zhao2024drivedreamer}, we further extend it into 12 Hz data. Specifically, we utilize the interpolated 12 Hz annotations from ASAP~\cite{wang2022asap} and the LiDAR data from the NuScenes dataset to generate semantic occupancy labels at 12 Hz. 
We begin the generation of the 12 Hz occupancy labels by concatenating the LiDAR points from the entire scene, following the method outlined in~\cite{wei2023surroundocc}. Next, we reconstruct the mesh of the driving scene using the NKSR algorithm~\cite{huang2023nksr}. We then extract the vertices of the mesh and assign semantic labels to these vertices based on the LiDARseg tags from NuScenes~\cite{caesar2020nuscenes} and the 12 Hz annotation information from ASAP~\cite{wang2022asap}. Finally, we convert the mesh vertices, along with their semantic labels, into semantic occupancy data.\looseness=-1

\section{Evaluation Metrics} 

To evaluate the fidelity of occupancy generation, we adopt the F3D metric and Maximum Mean Discrepancy (MMD) as~\cite{liu2023pyramid}. F3D represents a three-dimensional adaptation of the two-dimensional Fréchet Inception Distance (FID), utilizing a pre-trained occupancy auto-encoder. The MMD is calculated within the feature space of this same auto-encoder. Given that BEV layouts are used for generation, Mean Intersection over Union (mIoU) serves as the metric for evaluating the accuracy of the generated outputs. For occupancy forecasting, we report both IoU and mIoU, aligning with previous works \cite{zheng2023occworld}.
For video generation assessment, we utilize the FID and Frechet Video Distance (FVD) to evaluate the quality of the generated content, following previous works~\cite{heusel2017gans,unterthiner2018towards}. For the evaluation of LiDAR generation, we apply the Jensen-Shannon Divergence (JSD) and MMD following LiDARDM~\cite{zyrianov2024lidardm}.\looseness=-1

For the downstream perception task of Semantic Occupancy Prediction (SOP), we adopt mIoU and IoU as evaluation metrics, following previous research \cite{cao2022monoscene, li2024time}. For BEV segmentation, we use CVT~\cite{zhou2022cross} as the baseline and assess the mIoU for road and vehicle classes, following the approach in~\cite{gao2023magicdrive}. For 3D object detection, we employ BEVFusion~\cite{liu2023bevfusion} as the baseline model and measure performance using mean Average Precision (mAP) and the NuScenes Detection Score (NDS), as employed in~\cite{gao2023magicdrive}.\looseness=-1

\section{Model Setup}

The UniScene framework undergoes a two-stage training process implemented with PyTorch on NVIDIA A100 GPUs. Initially, the occupancy generative models are trained using ground-truth labels. 
Subsequently, the occupancy generative model is fixed to generate occupancy grids from the BEV maps, while the video and LiDAR generation models are jointly trained with occupancy-based conditions.\looseness=-1

The occupancy generation model is first trained with a batch size of 16 and an AdamW optimizer. The number of occupancy frames is fixed at 8 during training. For Occupancy VAE, the learnable class embedding $C'$ is set to 8 following ~\cite{zheng2023occworld}. 
The learning rate is set to $1 \times 10^{-3}$ over 200 epochs. For Occupancy DiT, the diffusion transformer model is based on ~\cite{peebles2023scalable}, with a learning rate of $1 \times 10^{-4}$ over 600 epochs.\looseness=-1

In the second stage, the video generation model and the LiDAR generation model are jointly trained for 90 epochs with a total batch size of 96 and an initial learning rate of $1 \times 10^{-5}$.
The video generation model is initialized with pre-trained weights from SVD~\cite{blattmann2023stable} with a fixed video VAE.
The image resolution for training is set to 256×512 following~\cite{wen2023panacea}, and the number of frames is fixed at 8. Note that we apply the roll-out sampling strategy following~\cite{gao2024vista} in the sampling process to enable long-term generation.
The LiDAR generation model is trained from scratch.
The AdamW optimizer is leveraged during the training.\looseness=-1

\begin{figure}[!t]
    \centering
\includegraphics[width=1.00\linewidth]{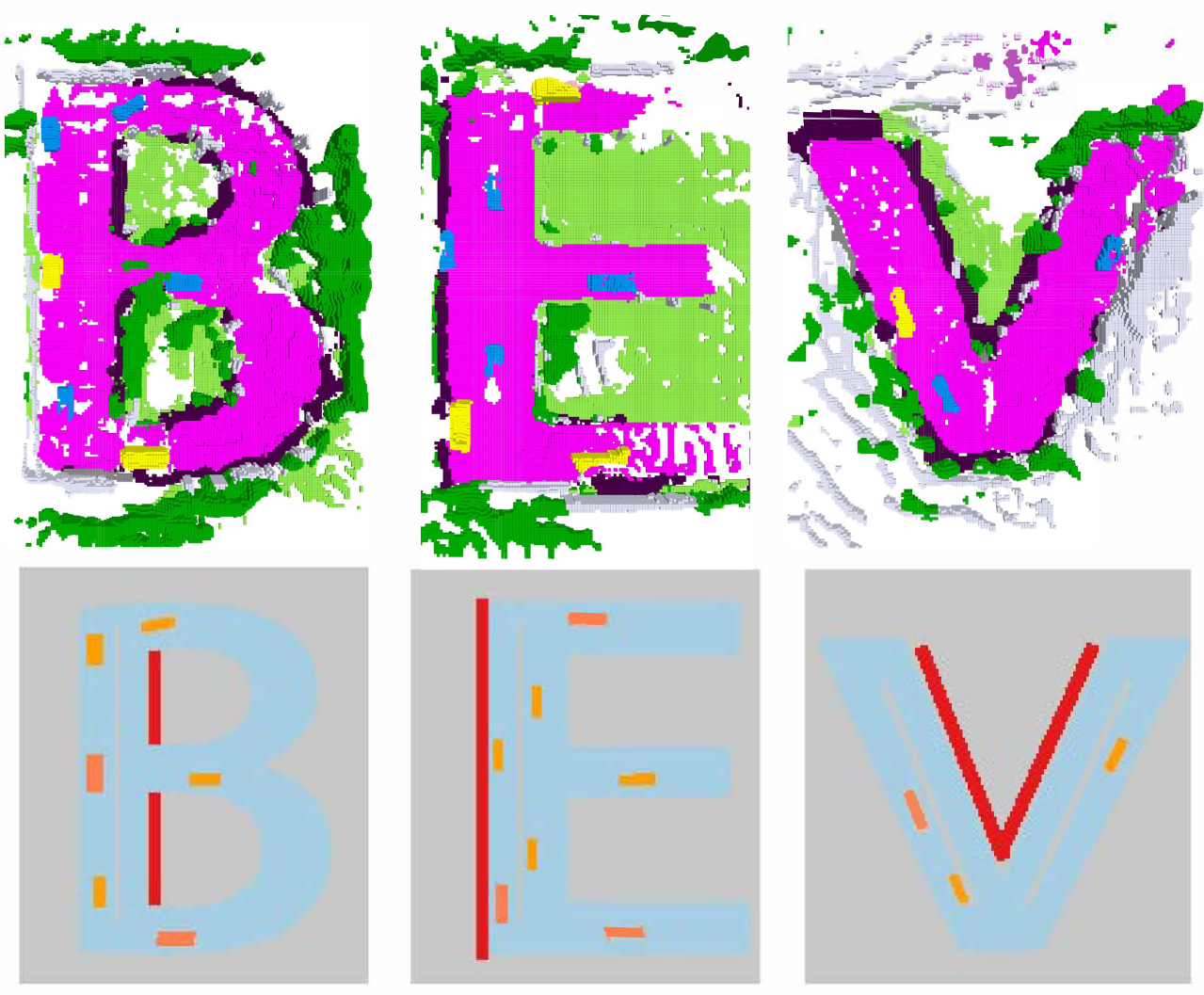}
    \caption{Visualization of Out-of-Distribution (OOD) occupancy generation, showcasing the model's ability to generalize and produce realistic occupancy outputs in unseen scenarios. The top row displays the generated 3D occupancy grids, while the bottom row presents the corresponding BEV layouts, highlighting the preservation of structural and semantic consistency in OOD cases.}
    \label{fig_ood_occ}
\end{figure}

\begin{figure}[!t]
    \centering
\includegraphics[width=1.00\linewidth]{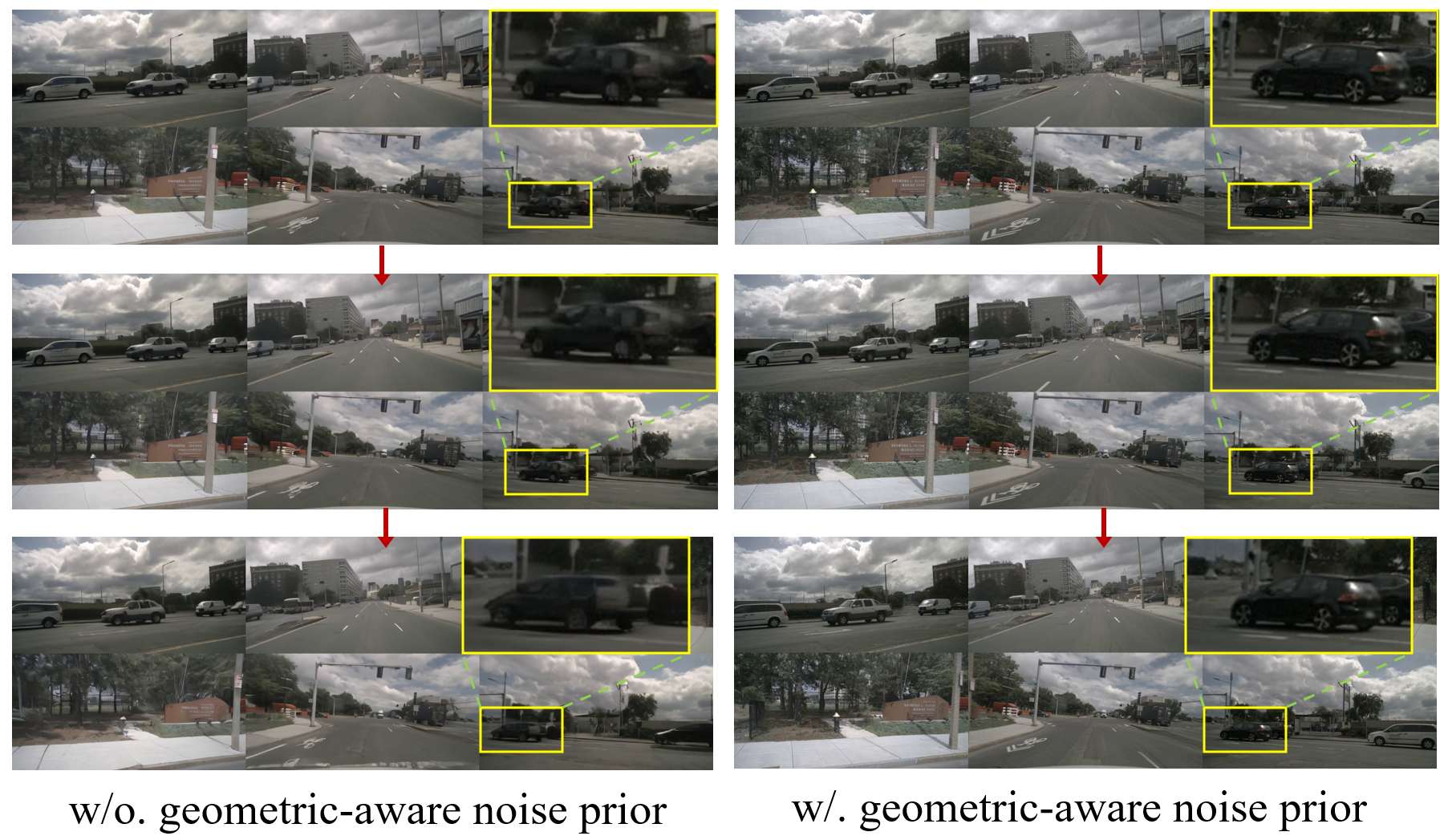}
    \caption{Visualization of the effect on geometric aware noise prior.
    Our method injects dense
appearance priors and incorporates explicit geometric awareness in the sampling process, resulting in high-fidelity and consistent moving cars across different frames.}
    \label{fig_noise_prior}
\end{figure}

\begin{figure*}[!t]
    \centering
\includegraphics[width=0.92\linewidth]{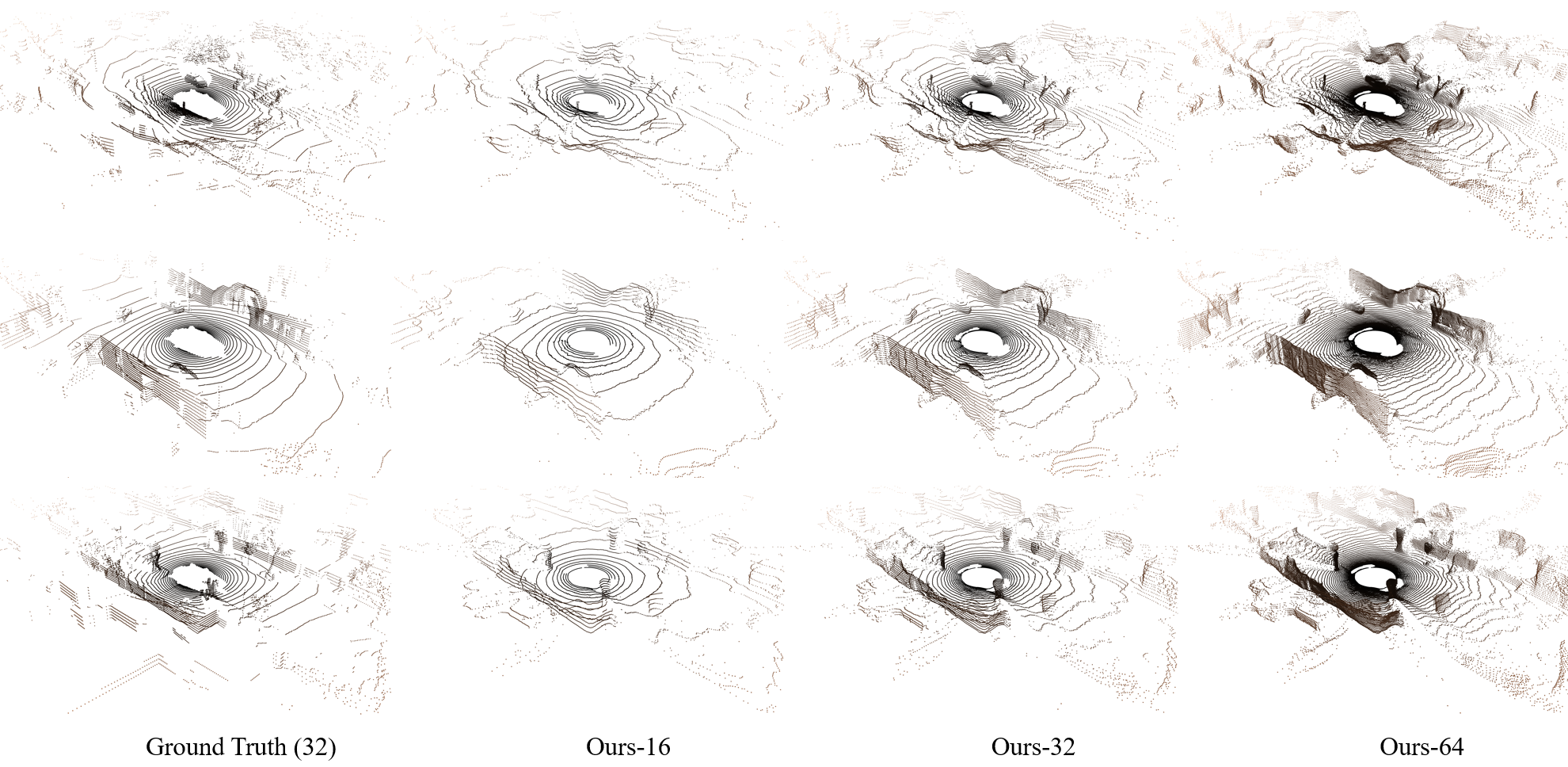}
    \caption{Visualization of LiDAR beam scanning patterns across different configurations. The ground truth utilizes a 32-beam LiDAR setup, while the proposed methods ('Ours-16', 'Ours-32', and 'Ours-64') demonstrate varying levels of beam density. The comparison highlights the model's ability to simulate realistic LiDAR patterns and preserve scene geometry across different beam resolutions.}
    \label{fig_lidar_ray}
\end{figure*}

\begin{figure*}[!t]
    \centering
    \includegraphics[width=0.98\linewidth]{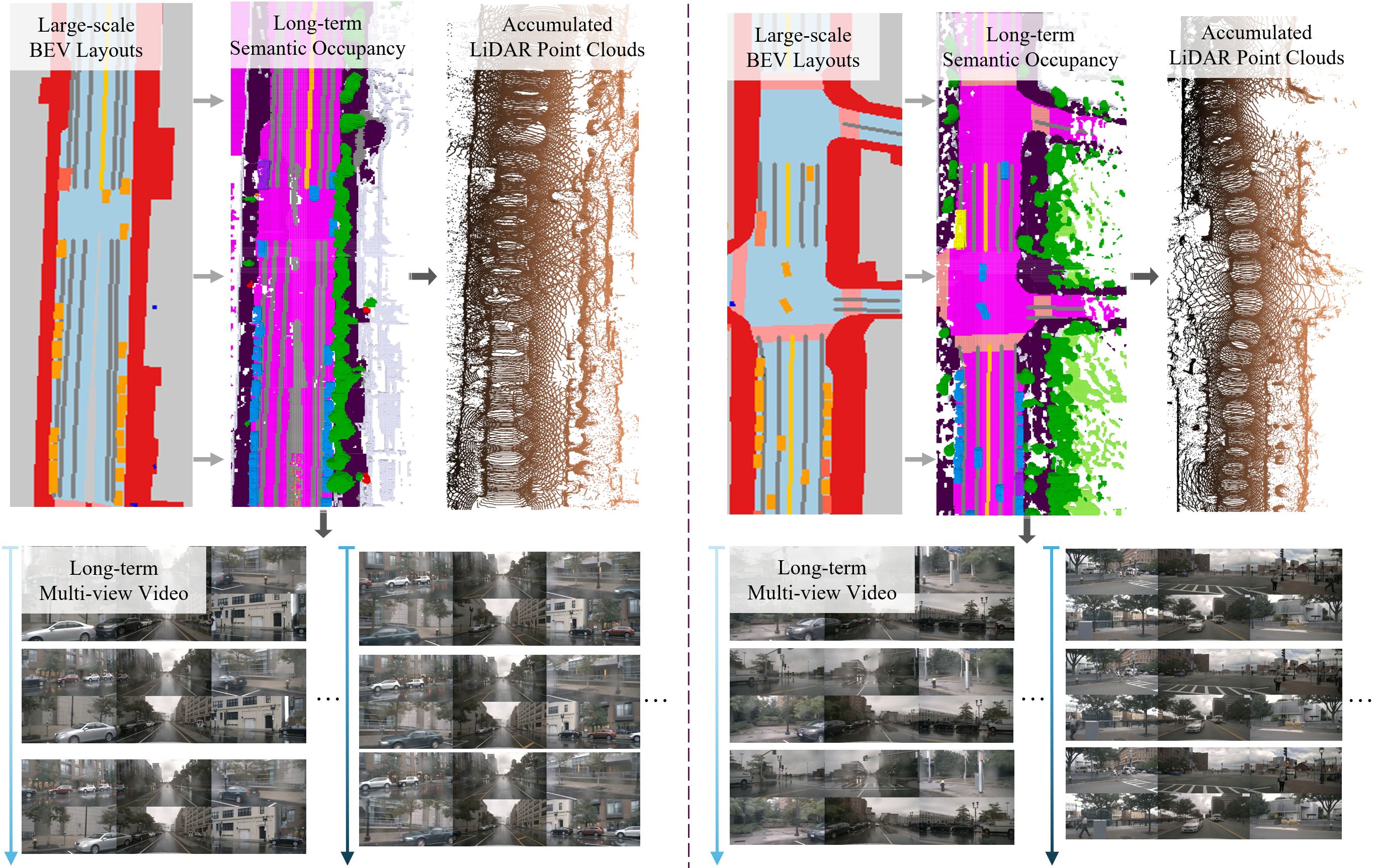}
    \caption{Visualization of large-scale coherent generation, where a large-scale BEV layout serves as the input. 
The long-term semantic occupancy from the given BEV layouts is first produced, which subsequently guides the generation of LiDAR point clouds and multi-view videos.
The results demonstrate the model's ability to produce temporally and spatially consistent outputs in large-scale environments.}
    \label{fig_lar}
\end{figure*}

\begin{figure*}[!t]
    \centering
\includegraphics[width=0.95\linewidth]{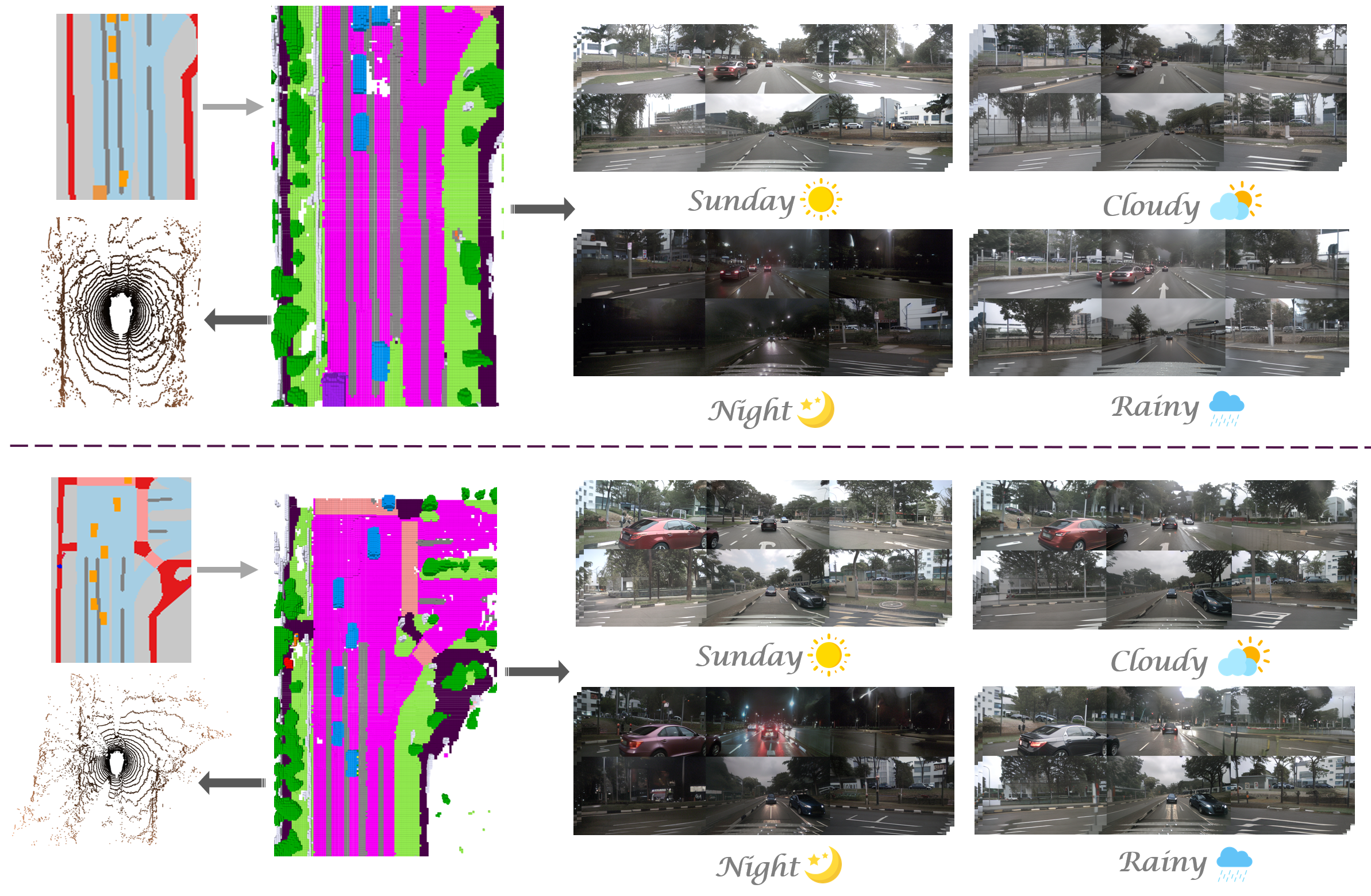}
    \caption{Visualization of controllable video generation with diverse attributes. The pipeline demonstrates the ability to generate videos conditioned on BEV layouts, with control over various attributes such as weather (sunny, cloudy, rainy) and time of day (day, night). The top and bottom rows showcase different input configurations, resulting in realistic video outputs with the desired attribute variations.}
    \label{fig_wea}
\end{figure*}

\begin{figure*}[!ht]
  \begin{scriptsize}
    \centering
    \includegraphics[width=0.9\linewidth]{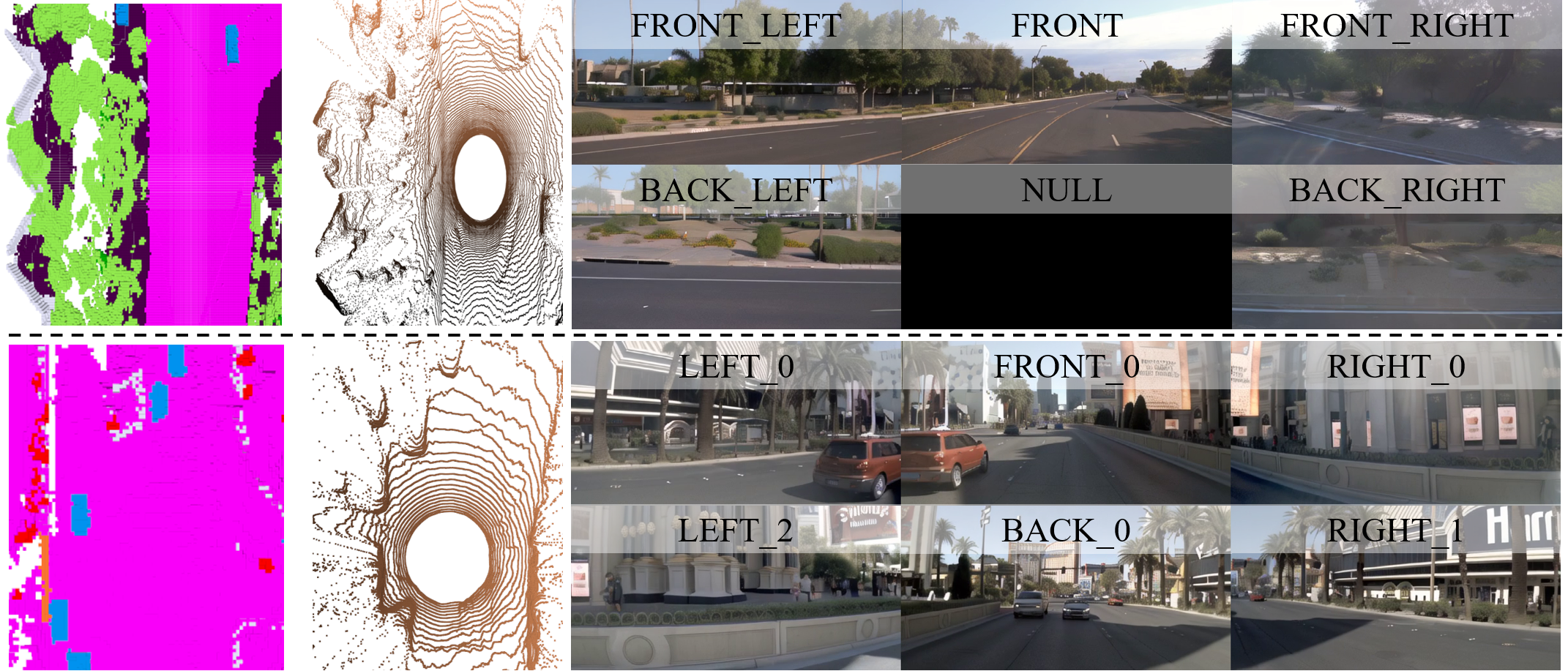}
    \caption{Generalization on the Waymo (upper) and nuPlan (lower) datasets. Due to unavailability, the back view on the Waymo dataset is set to null.\looseness=-1}
    \label{reb3}
     \end{scriptsize}
\end{figure*}

\section{Comparison Baselines}
The occupancy generation method is compared with OccLlama \cite{wei2024occllama}, OccWorld \cite{zheng2023occworld}, and OccSora \cite{wang2024occsora}. OccWorld relies on historical occupancy data and the trajectory of the ego vehicle to predict future occupancy. Building upon OccWorld, OccLlama incorporates a Large Language Model (LLM) to interpret the semantic scene, thereby enhancing forecasting accuracy.\looseness=-1 

We compare our video generation results with other image and video generation methods, including
BEVGen~\cite{swerdlow2024streetview},
BEVControl~\cite{yang2023bevcontrol},
DriveGAN~\cite{kim2021drivegan},
DriveDreamer~\cite{zhao2024drivedreamer},
WoVoGen~\cite{lu2023wovogen},
Panacea~\cite{wen2023panacea},
MagicDrive~\cite{gao2023magicdrive},
Drive-WM~\cite{wang2023driving},
Vista~\cite{gao2024vista}. As mentioned in Sec.4.1 of the main paper, the initial model of Vista~\cite{gao2024vista} only supports single-view video generation. Thus, we implement the multi-view variant of Vista$^*$ with spatial-temporal attention following~\cite{wu2023tune,gao2023magicdrive} for a fair comparison.\looseness=-1

The LiDAR generation results are compared with Open3D \cite{zhou2018open3d} and LiDARDM \cite{zyrianov2024lidardm}. 
Open3D employs a hard ray-casting function to generate LiDAR point clouds from occupancy grids. This process involves converting the occupancy grid into a mesh, where each occupied voxel is represented as a cube. The ray depth is determined by calculating the intersection between the ray and the nearest triangular face of the mesh. Note that this method does not provide intensity values or ray drop probabilities.
LiDARDM begins the generation process by aggregating multi-frame point clouds and removing dynamic objects using 3D object detection labels. The remaining data is used to reconstruct a mesh via the NKSR algorithm~\cite{huang2023nksr}. A mesh diffusion model is trained using this mesh as a 3D representation and the BEV layout as a condition. During inference, a mesh is sampled from the diffusion model based on a BEV layout map. Mesh models of dynamic objects, selected from predefined assets, are then inserted into the generated mesh according to the detection labels of the corresponding frame. Hard ray casting is subsequently applied to obtain the point cloud. Similar to Open3D, LiDARDM does not generate intensity values and requires an additional network to predict ray drop probabilities.\looseness=-1

\begin{figure*}[!t]
    \centering
    \includegraphics[height=0.94\textheight]{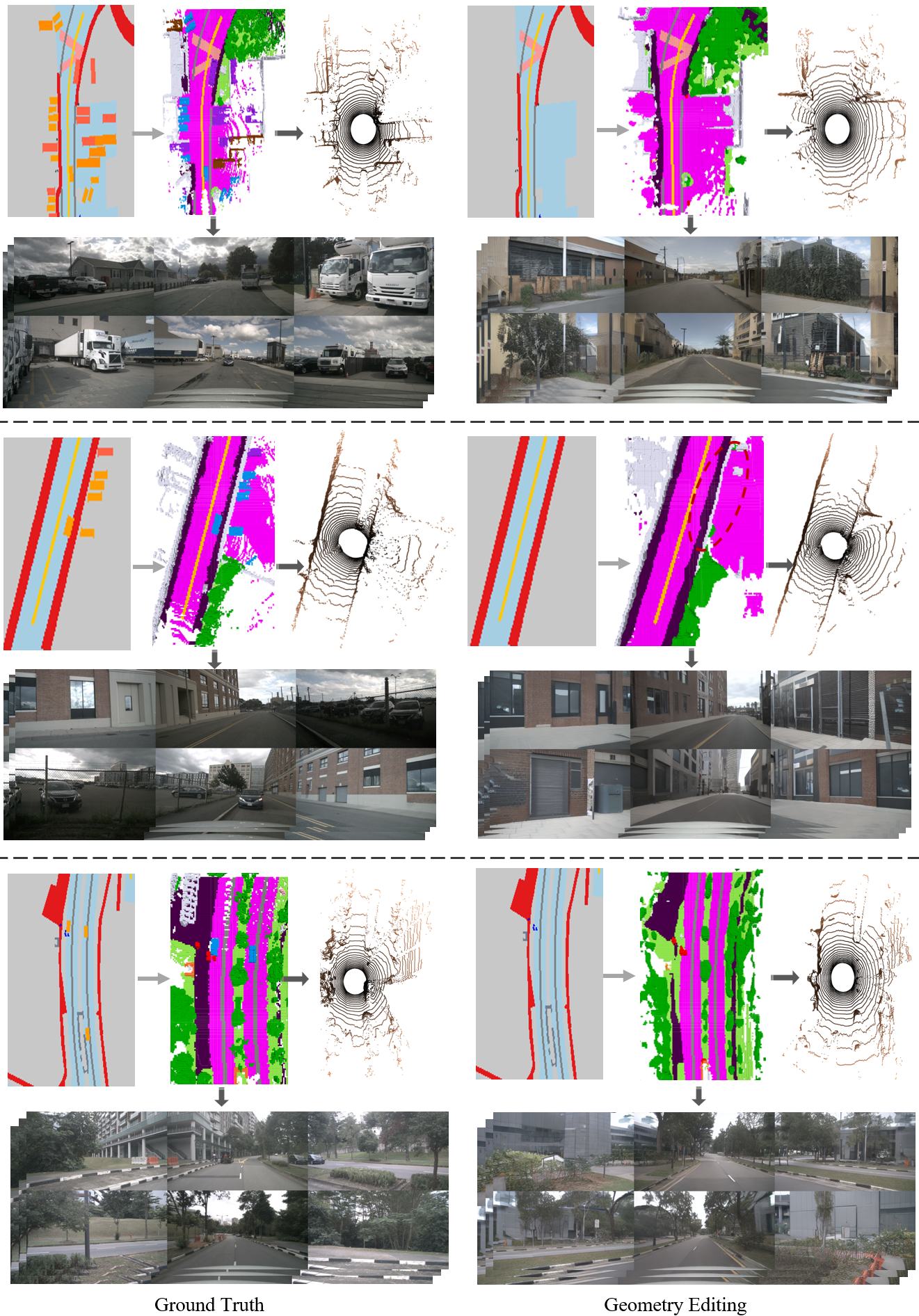}
    \caption{Visualization of controllable generation through geometry editing, demonstrating the ability to manipulate BEV layouts to alter scene geometry. The process begins with modified BEV layouts, followed by the generation of updated semantic occupancy and LiDAR point clouds. These results in multi-view video outputs that reflect the geometric changes, showcasing the model's flexibility in producing consistent and realistic outputs based on edited scene geometries.}
    \label{fig_geo_edit}
\end{figure*}

\section{More Visualization Results}

\noindent\textbf{Out-of-Distribution Occupancy Generation.}
To further illustrate the strong generalization capability and controllability of our occupancy generation model, we present the Out-of-Distribution (OOD) occupancy generation results, as shown in \cref{fig_ood_occ}. These visualizations demonstrate the model’s ability to generate realistic occupancy outputs in previously unseen scenarios, while maintaining structural and semantic consistency in OOD cases.

\noindent\textbf{Generalization on other Autonomous Driving Datasets.}
As shown in Fig.~\ref{reb3}, we evaluate the generalization ability of our model on the Waymo and nuPlan datastes. Given scene layouts, our model can be directly transferred to different datasets with conditional reference images. 
Note that due to unavailability, the back view on the Waymo dataset is set to null.\looseness=-1

\noindent\textbf{Effect on Geometric Aware Noise Prior.}
As mentioned in Sec.3.2 of the main paper, the Geometric-aware Noise Prior strategy injects dense
appearance priors and incorporates explicit geometric awareness in the sampling process of video generation.
The Visualization results are illustrated in Fig.~\ref{fig_noise_prior}. 
We can see that the video generation quality is obviously improved, resulting in high-fidelity and consistent moving cars across different frames.

\noindent\textbf{Diverse LiDAR Beam Scanning Patterns.}
Owing to the flexibility of our ray-based LiDAR heads, the scanning pattern of the LiDAR beam can be freely configured. This capability facilitates the generation of LiDAR point clouds with various scanning patterns without necessitating retraining. As illustrated in Fig.~\ref{fig_lidar_ray}, whereas the ground truth LiDAR beam is constrained to 32, our method is capable of generalizing to 16, 32, and 64 LiDAR beams.
The visualization underscores the model's capability to simulate realistic LiDAR patterns while preserving scene geometry across various beam resolutions.\looseness=-1

\begin{figure*}[!t]
    \centering
    \includegraphics[height=0.93\textheight]{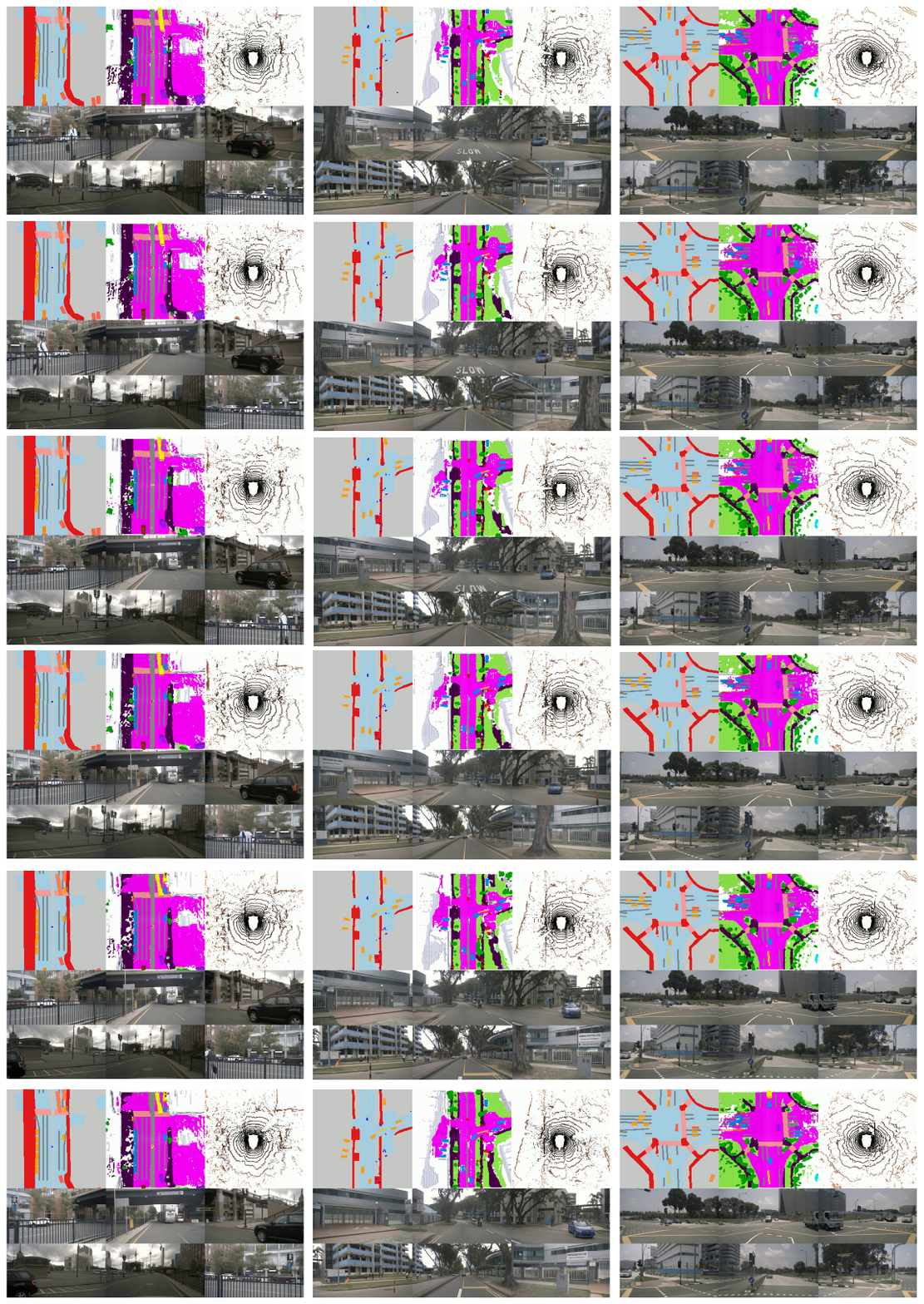}
    \caption{More visualization of unified generation of semantic occupancy, LiDAR point clouds, and multi-view videos.}
    \label{fig_ug1}
\end{figure*}

\begin{figure*}[!t]
    \centering
    \includegraphics[height=0.95\textheight]{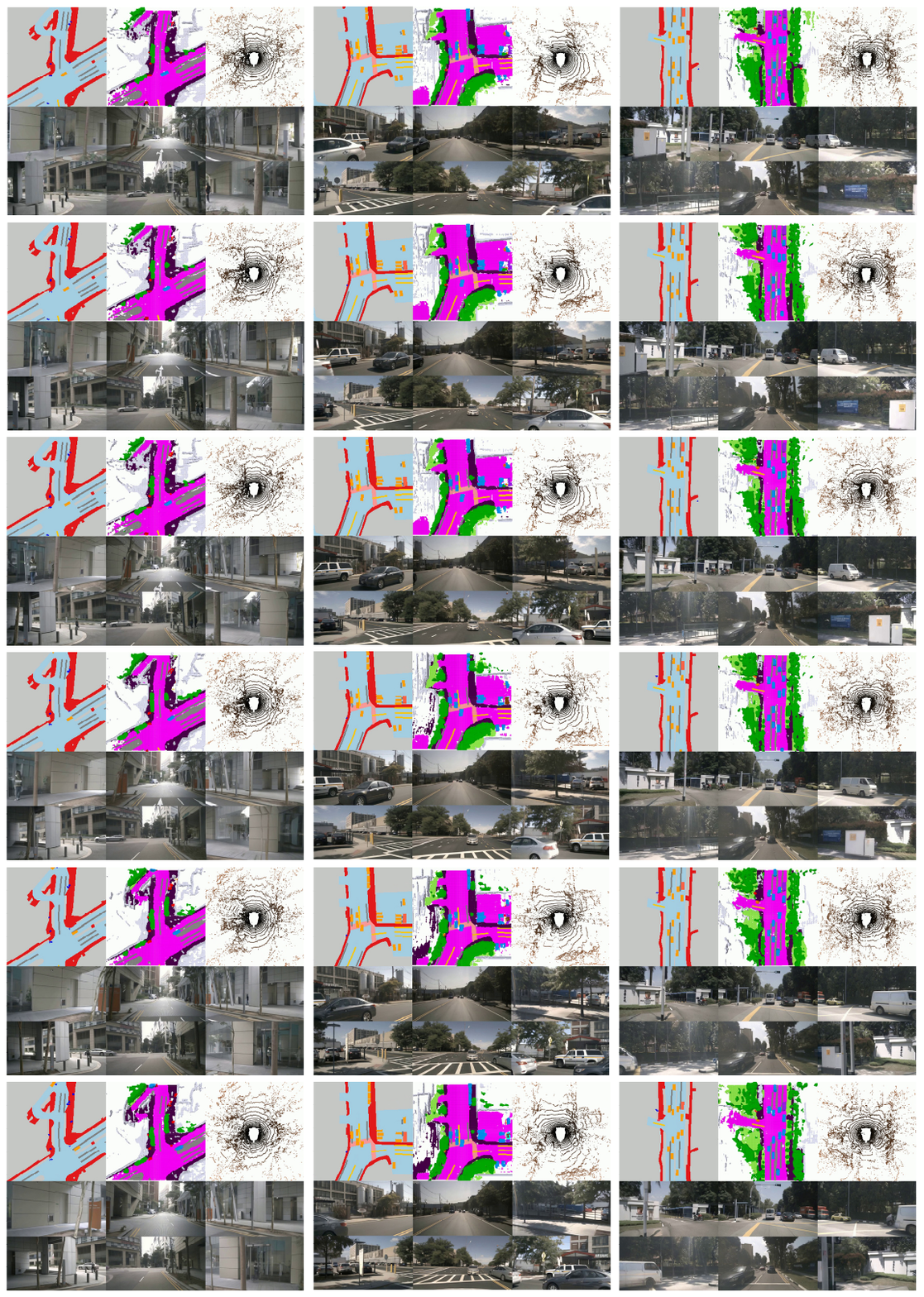}
    \caption{More visualization of unified generation of semantic occupancy, LiDAR point clouds, and multi-view videos.}
    \label{fig_ug2}
\end{figure*}

\begin{figure*}[!t]
    \centering
    \includegraphics[height=0.95\textheight]{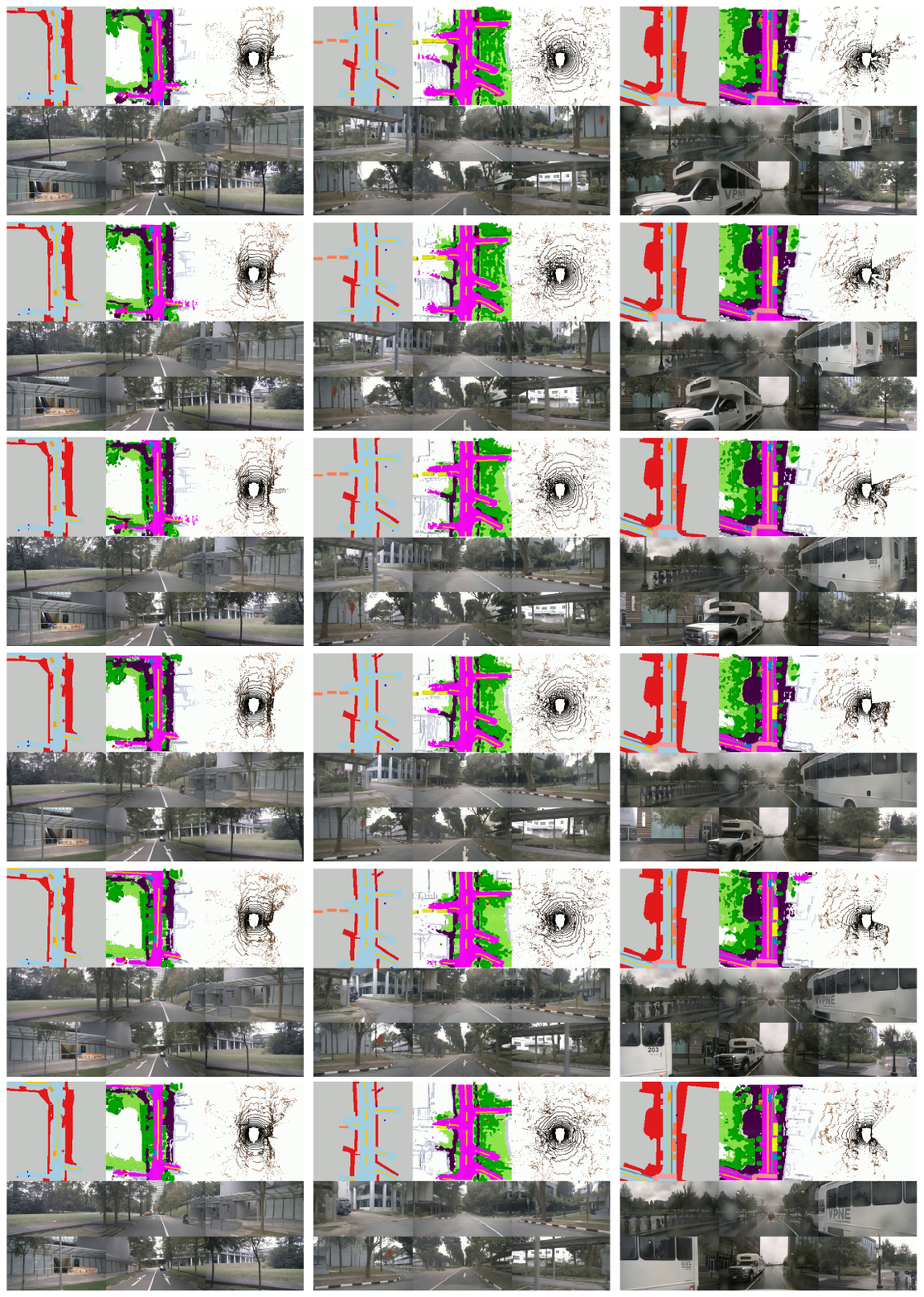}
    \caption{More visualization of unified generation of semantic occupancy, LiDAR point clouds, and multi-view videos.}
    \label{fig_ug3}
\end{figure*}

\noindent\textbf{Versatile Generation Ability of UniScene.}
To further demonstrate the versatile generation capabilities of our proposed UniScene, we apply it across various scenarios. 
Figure~\ref{fig_lar} illustrates UniScene's capacity to generate large-scale, coherent scenes, underscoring the model’s ability to produce outputs that are temporally and spatially consistent in extensive environments.
The controllable generation of attribute-diverse videos is showcased in Figure~\ref{fig_wea}, highlighting realistic video outputs with desired attribute variations for different input configurations. 
Additionally, the application of UniScene to scene editing is depicted in Figure~\ref{fig_geo_edit}, demonstrating the model’s flexibility in generating consistent and realistic outcomes based on edited scene geometries. 
The results presented in Figures~\ref{fig_ug1},~\ref{fig_ug2}, and~\ref{fig_ug3} further exemplify UniScene's capability to jointly produce high-quality semantic occupancy, video, and LiDAR data, all while maintaining temporal consistency.\looseness=-1

\newpage
\clearpage

{
    \small
    \bibliographystyle{ieeenat_fullname}
    \bibliography{main}
}

\end{document}